\documentclass[10pt,twocolumn,letterpaper]{article}

\usepackage[pagenumbers]{cvpr} 

\usepackage{tcolorbox}
\tcbuselibrary{skins,breakable}
\usepackage{enumitem}

\newtcolorbox{promptbox}[1]{%
  enhanced,
  breakable,
  title={#1},
  title after break={#1\space (continued)},
  colback=black!1,
  colframe=black!30,
  colbacktitle=black!7,
  coltitle=black,
  fonttitle=\normalfont\normalsize\bfseries,
  fontupper=\normalfont\normalsize,
  boxrule=0.5pt,
  arc=2pt,
  boxsep=0pt,
  left=8pt,
  right=8pt,
  top=7pt,
  bottom=7pt,
  toptitle=5pt,
  bottomtitle=5pt,
  before skip=10pt,
  after skip=10pt,
  before upper={%
    \raggedright
    \setlength{\parindent}{0pt}%
    \setlength{\parskip}{0.5em}%
  },
}

\newlist{promptitems}{itemize}{1}
\setlist[promptitems]{%
  label=\textbullet,
  leftmargin=1.3em,
  topsep=0.3em,
  itemsep=0.3em,
  parsep=0pt,
  partopsep=0pt
}

\definecolor{cvprblue}{rgb}{0.21,0.49,0.74}
\usepackage[pagebackref,breaklinks,colorlinks,allcolors=cvprblue]{hyperref}
\usepackage{etoc}
\etocdepthtag.toc{main}

\def\confName{CVPR}
\def\confYear{2027}

\usepackage{xspace}
\newcommand{\ours}{PREP\xspace}

\title{\ours: Predictive Polymorphic Memory for Long-form Egocentric Videos}

\newcommand{\aff}[2]{\normalfont\normalsize\textsuperscript{#1}#2}
\author{%
  Lorenzo Venturoli\textsuperscript{1}, Nikola Popovic\textsuperscript{1}, Alessio Tonioni\textsuperscript{2}, Yongqin Xian\textsuperscript{2},\\
  Luc Van Gool\textsuperscript{1}, Federico Tombari\textsuperscript{2}, Danda Pani Paudel\textsuperscript{1}\\[0.6em]
  \aff{1}{Institute for Computer Science, Artificial Intelligence and Technology (INSAIT)}\\
  \aff{2}{Google}\\[0.4em]
  \normalfont\small\texttt{lorenzo.venturoli@insait.ai}
}
\date{}

\begin{document}

\twocolumn[{%
    \begin{@twocolumnfalse}
        \maketitle

        \begin{minipage}{\textwidth}
            \centering
            \includegraphics[width=\linewidth]{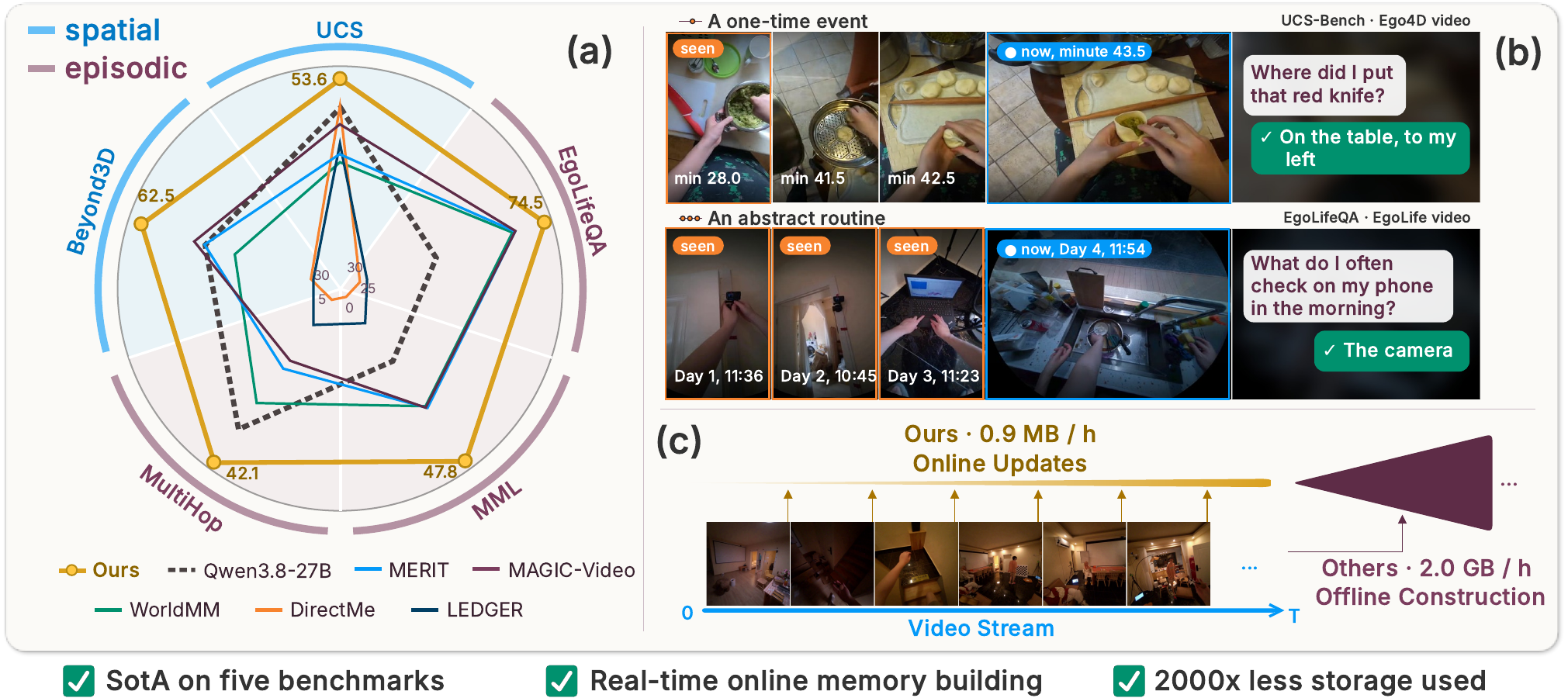}
            \captionof{figure}{\textbf{\ours memory at a glance.}
            \textbf{(a)} \ours achieves state-of-the-art performance across five episodic and spatial egocentric benchmarks, demonstrating broad generalization.
            \textbf{(b)} Qualitative examples with video frames and question-answer pairs.
            \textbf{(c)} Lowest memory storage (MB/hour) among state-of-the-art methods, while uniquely supporting native online video processing in real-time.}

            \label{fig:teaser}
        \end{minipage}
        \vspace{1em}
    \end{@twocolumnfalse}
}]

\begin{abstract}
%

Long-form memory is crucial for egocentric video understanding and its applications in augmented reality and robotics, since questions often hinge on what happened, who was involved, when, and where, minutes or hours earlier. However, existing memory methods are too specialized. Episodic methods lose visual and spatial detail, while spatial methods ignore events and time. None holistically outperforms a general-purpose vision-language model (VLM), and most cannot process streaming video or require hundreds of megabytes per hour. We propose \ours, a streaming memory framework with two branches. The polymorphic branch organizes memory as an evolving tree of concepts, from one-time events to abstract routines. The predictive branch poses challenges about incoming frames, answers them from memory, and forwards the results to the polymorphic branch, which incorporates missing information or reinforces existing knowledge. Built entirely from training-free VLM agents, \ours achieves state-of-the-art results on five episodic and spatial egocentric benchmarks, runs on streaming video in real time, and uses under 1\,MB/hour of storage. 
Our analysis shows that \ours retrieves correct evidence more often than competing methods, while nodes summarizing recurring patterns improve accuracy and efficiency. 
Project page: \url{prep.insait.ai}.

\end{abstract}


\section{Introduction}

%
Egocentric videos provide a continuous first-person view of a human or embodied agent interacting with the environment, making them central to emerging applications in augmented and mixed reality (AR/MR) and robotics~\cite{Grauman_2022_CVPR, yang2025egolife, openEQA}. These applications require methods that analyze such streams and answer questions in real time, enabling context-aware scene understanding and action recommendations. Unlike conventional third-person videos, egocentric videos are often long-form, contain frequent occlusions, and involve substantial camera motion~\cite{Grauman_2022_CVPR, perrett2025hdepic, Grauman_2024_CVPR}. The long-form nature is especially consequential since answering a question often depends on observations from minutes or hours earlier, such as what happened, who was involved, when, and where. Long-form egocentric understanding, and the memory that supports it, is therefore the cornerstone of all egocentric applications~\cite{yang2025egolife, chen2026multimodallifelongunderstandingdataset, chen2025grounded, Grauman_2022_CVPR}.

To assess the current landscape, we evaluate a wide range of methods on the two most prominent types of egocentric memory benchmarks: episodic and spatial. Episodic benchmarks (EgoLifeQA~\cite{yang2025egolife}, MM-Lifelong~\cite{chen2026multimodallifelongunderstandingdataset}, MultiHop-QA~\cite{chen2025grounded}) test recall of what happened, who was involved, and when, often requiring long-term, multi-hop retrieval. Spatial benchmarks (UCS-Bench~\cite{wang2026keep}, Beyond3D~\cite{ma2026beyond3d}) test recall of where objects are and how they relate in 3D, requiring spatial reasoning. We find that specialized egocentric methods excel on their target tasks but degrade substantially elsewhere. Episodic memory methods~\cite{wang2024lifelongmemoryleveragingllmsanswering, zhang-etal-2024-simple, choi2026keep} store captions of fixed time slices, losing visual and spatial detail and breaking down on open-ended questions. Spatial memory methods~\cite{wang2026keep, chaudhari2026lookbackunderstandingpersistence, 3dmem} instead store object locations but not events, people, or time, and thus fail on event- and time-centric tasks. In fact, none of these methods holistically outperforms the general-purpose vision-language model Qwen-VL~\cite{bai2025qwen3vltechnicalreport}. Specifically, every method is worse than Qwen on at least \textit{two out of five} benchmarks.

Beyond accuracy, existing methods share two practical limitations: most (i) cannot process streaming video in real time and (ii) require large amounts of storage. Although egocentric video naturally arrives as a stream in real-world applications~\cite{videollm,wang2026keep}, most methods either build their memory offline from the entire video~\cite{li2026magic, yeo2026worldmm} or consume only a fixed number of sub-sampled frames~\cite{chen2026multimodallifelongunderstandingdataset}. Moreover, most store past frames as visual evidence, requiring hundreds of megabytes per hour of video~\cite{3dmem, yeo2026worldmm}. Thus, no existing approach handles the full range of egocentric tasks while operating on streaming video with a compact memory. To fill this gap, we propose \ours{}, a holistic memory framework that achieves state-of-the-art performance on all five episodic and spatial egocentric benchmarks, runs on streaming video in real time, and uses less than one megabyte of storage per hour of video (Figure~\ref{fig:teaser}).

\begin{figure}[t]
    \centering
    \includegraphics[width=\linewidth]{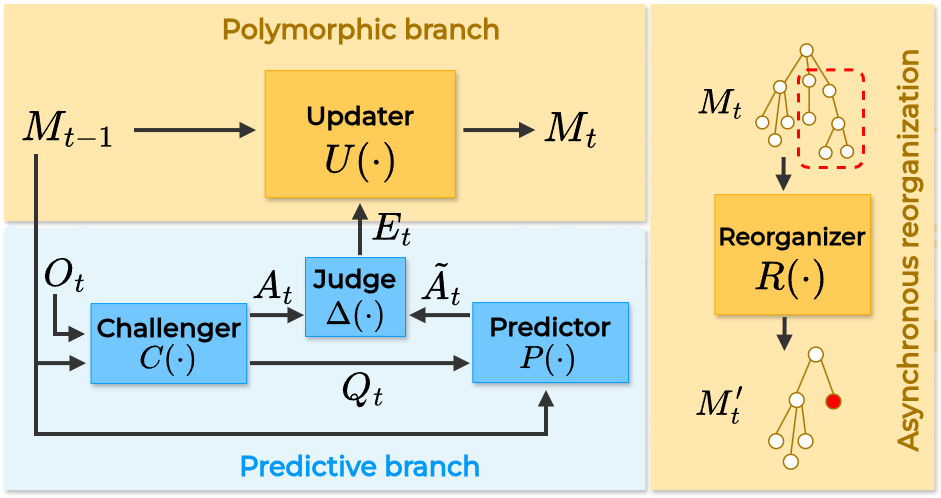}
    \vspace{-5mm}
    \caption{\textbf{\ours framework.}
    The predictive branch probes memory with challenges from video observations. The polymorphic branch incorporates missing evidence, reinforces correctly predicted information, and reorganizes memory into abstractions.}
    \label{fig:diagram}
    \vspace{-5mm}
\end{figure}

\ours consists of a polymorphic memory branch, which determines \emph{how} knowledge is stored, and a predictive branch, which determines \emph{what} is stored (Figure~\ref{fig:diagram}). 
The polymorphic branch builds on the assumption that knowledge from videos comes in many forms and keeps evolving, 
so \emph{the memory must flexibly organize diverse knowledge}~\cite{AMem, memtree}. 
For example, ``Where did I put the red knife?'' requires recalling a specific one-time event, whereas ``What do I check on my phone in the morning?'' requires inferring a routine that emerges only over many mornings. Repeated observations start as separate events and gradually become an abstract concept. The memory is therefore organized as a tree of concept nodes, with one-time events and fine details at lower levels, routines and abstract concepts at higher levels, and separate subtrees for different types of knowledge.

The predictive branch builds on the assumption that \emph{a memory only needs to store new information that it cannot already predict}~\cite{schmidhuber1992learning,gorlo2026worthrememberingsurprisegatedrobot,NEURIPS2025_a4ca07aa}. Therefore, it continuously tests the memory against the incoming stream. A challenger analyzes the most recent frame window and proposes question-answer pairs about what was just observed, which we call challenges. A predictor answers each challenge from the current memory, and a judge checks the answers.  
Correctly answered challenges reinforce retrieved memory nodes with new observation timestamps, without duplicating stored evidence. Failed challenges instead trigger updates to incorporate missing information. The reorganizer then consolidates related nodes into higher-level abstractions.
For instance, turning morning phone checks into a routine. We implement each sub-module as a general-purpose VLM agent without any re-training whatsoever. During inference, an agent reads the memory to answer the question.

Beyond overall performance, our experiments reveal three key findings. First, \ours retrieves relevant evidence more reliably than competing memory methods, particularly on week-long videos, and maintains its accuracy even when questions are delayed by several days, unlike leading baselines. Second, predictive filtering and polymorphic reorganization are complementary. Removing both reduces retrieval accuracy by 19.9\%, while their combination reduces stored memory nodes by 29\% and improves retrieval efficiency by 32\%.  Third, despite having no imposed hierarchy structure, \ours spontaneously develops a hierarchical memory structure, progressively consolidating observations into higher-level abstractions that capture recurring patterns across days and facilitate efficient retrieval.

Our main contributions are:
\begin{itemize}
    \item \ours, a real-time streaming memory framework with predictive filtering and polymorphic organization.
    \item SotA results on five episodic and spatial benchmarks, with real-time streaming and $<$1\,MB per video hour.
    \item An analysis of retrieval gains from both branches and emergent memory hierarchies without explicit design.
\end{itemize}

\section{Related Work}

\noindent\textbf{Egocentric Video Understanding.}
Egocentric video understanding has expanded from recognizing everyday activities and skilled interactions~\cite{epickitchens,Grauman_2022_CVPR,Grauman_2024_CVPR,perrett2025hdepic} to learning language-aligned representations~\cite{lavila} and answering causal, first-person, and temporally grounded questions~\cite{egotaskqa,egothink,egoschema,chen2025grounded}.
Recent episodic benchmarks extend this setting to days or weeks of personal experience~\cite{yang2025egolife,chen2026multimodallifelongunderstandingdataset,wang2026egomemreasonmemorydrivenreasoningbenchmark}, while spatial benchmarks require persistent object and viewpoint reasoning~\cite{openEQA,wang2026keep,ma2026beyond3d}.
To accommodate longer histories, video-language models compress visual context into memory banks~\cite{moviechat,malmm,rewind}, propagate information recurrently~\cite{videostreaming,videollamb}, or maintain memories for online interaction~\cite{videollm,flashvstream,streamchat,streamforest}.
However, existing memory methods are tailored to specific tasks and benchmarks, limiting their generalization to other settings. We propose a unified memory framework that achieves top performance on diverse benchmarks spanning episodic and spatial reasoning.

\noindent\textbf{Episodic Memory.}
Caption-based methods summarize video~\cite{wang2024lifelongmemoryleveragingllmsanswering,zhang-etal-2024-simple} for document or semantic retrieval~\cite{drvideo,choi2026keep}. Other approaches gather evidence iteratively~\cite{fan2024videoagent,wang2024videoagent,rege2026agenticlongvideounderstanding}, enrich episodic records~\cite{long2025seeinglisteningrememberingreasoning,yeo2026worldmm,li2026magic,ren2026GEB}, organize evidence hierarchically~\cite{videorecap,wang2025videotreeadaptivetreebasedvideo,raptor}, revise memory dynamically~\cite{memtree,AMem,wang2026memforestefficientagentmemory,videomem}, prioritize surprising observations~\cite{schmidhuber1992learning,NEURIPS2025_a4ca07aa,gorlo2026worthrememberingsurprisegatedrobot, fountas2025emllm, ge2026streamingvideomodelremember, ferreira-etal-2026-episodic}, or store what a dialogue memory mispredicts~\cite{nemori2025}. Surprise and novelty signals can retain what is unexpected yet miss what the memory cannot answer. In contrast, \ours uses predictive challenges to expose exactly these gaps and retain what the memory fails to infer, regardless of knowledge type.

\noindent\textbf{Spatial Memory.}
Early episodic QA tested object and location recall~\cite{Datta_2022_CVPR,Barmann_2022_CVPR}. Spatial methods encode context~\cite{egoenv}, localize objects~\cite{egoloc}, and store queryable maps~\cite{conceptfusion}, relational graphs~\cite{conceptgraphs,Werby-RSS-24,graphEQA}, or representative views~\cite{3dmem}. Later methods model interactions~\cite{amego,Huang_2025_CVPR}, changing object locations~\cite{wang2026keep,chaudhari2026lookbackunderstandingpersistence}, and spatiotemporal relations~\cite{Gorlo_2026_CVPR}, while benchmarks probe spatial reasoning and out-of-sight persistence~\cite{openEQA,vsibench,ma2026beyond3d}. Spatially specialized memories can miss event histories and recurring patterns needed for broader reasoning. In contrast, \ours{} uses a flexible hierarchy to store events, object locations, patterns, and abstractions.



\noindent\textbf{Memory in Cognitive Science.}
Human memory pairs a short-lived store with a durable
one~\cite{atkinson1968human}, holds episodes and general
knowledge~\cite{tulving1972episodic}, and learns fast and
slow~\cite{mcclelland1995there,hinton1987using}. Early systems reorganized
memory as cases arrived~\cite{schank1980language,kolodner1983maintaining,gershman2017computational},
and prediction decides what to keep and when to
update~\cite{anderson1991reflections,reynolds2007computational,franklin2020structured,lu2022neural}. Recent vision and language memories adopt these principles~\cite{cheng2022xmem,gutierrez2025hipporag2,hema2025,humanmemory2026}.
They model each principle separately, with hand-built schemas, small trained networks, or fixed stores. \ours{} combines them in one system that needs neither and scales to week-long video.

\section{Method}
\label{sec:met}

\begin{figure*}[ht]
    \centering
    \includegraphics[width=\linewidth]{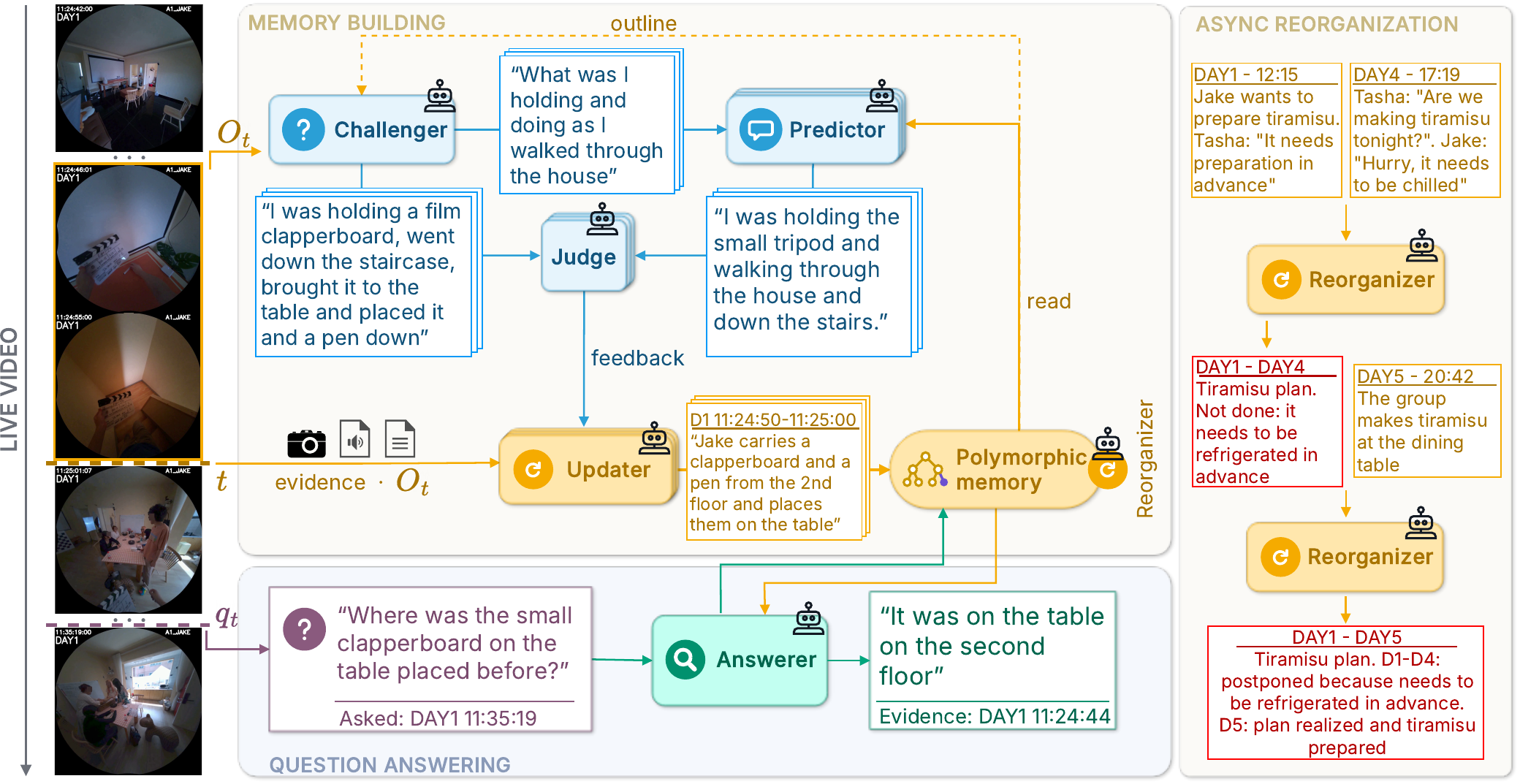}
    \vspace{-6mm}
    \caption{\textbf{Agentic architecture of \ours.} For each new video observation, the predictive branch uses challenger, predictor, and judge VLM agents to generate, answer, and evaluate challenges against existing memory. The updater incorporates missing evidence from failed challenges and reinforces existing memory from correct predictions, while the reorganizer consolidates recurring patterns into abstractions.}
    \vspace{-4mm}
    \label{fig:arch}
\end{figure*}

\subsection{Overview}

We introduce \ours, a memory-building framework for long-form videos, organized into two complementary branches.
The predictive branch (Sec.~\ref{sec:pred}) identifies missing information to store and existing information to reinforce. The polymorphic branch (Sec.~\ref{sec:poly}) determines how to store this evidence, allowing the memory to evolve by inserting new nodes and progressively merging related nodes into routines and abstractions. The resulting memory supports retrieval and visual question answering over long-form videos (Sec.~\ref{sec:qa}). Fig.~\ref{fig:diagram} shows our framework.


Our framework processes each new observation ${O_t = (V_t, S_t, C_t)}$ from the streaming video. Here, $V_t \in \mathbb{R}^{f_t \times H \times W \times 3}$ denotes the latest video clip of $f_t$ frames, while $S_t, C_t \in \mathcal{T}$ denote the corresponding speech transcript and video caption, with $\mathcal{T}$ representing the space of text sequences. Given $O_t$, the framework updates the existing memory as
\begin{equation}
    M_t = f(O_t, M_{t-1}).
\end{equation}
Our framework maintains memory $M_t$ as a rooted structure whose tree hierarchy emerges without a predefined schema:
\begin{equation}
\begin{aligned}
    M_t &= (\mathcal{N}_t, \mathcal{E}_t, r_t), \\
    \mathcal{N}_t &= \{(i, n_i) \mid i \in \mathcal{I}_t\}, \\
    \mathcal{E}_t &= \{(\pi_t(j), j) \mid j \in \mathcal{I}_t \setminus \{r_t\}\},
\end{aligned}
\end{equation}
where $\mathcal{I}_t = \{0, \ldots, |\mathcal{N}_t|-1\}$ is the set of unique non-negative integer node identifiers, each $n_i \in \mathcal{T}$ denotes the textual content of node $i$, $r_t \in \mathcal{I}_t$ is the root identifier, and $\pi_t(j)$ denotes the unique parent of node $j$ at the preceding tree level. The objective of \ours is to construct a compact memory $M_t$ that preserves the information needed to answer downstream questions $q$, such that ${p(y \mid q, M_t) = p(y \mid q, V_{\leq t})}$, where $V_{\leq t} = (V_1, \ldots, V_t)$ denotes all previously observed video clips.

\subsection{Predictive Branch}
\label{sec:pred}

The predictive branch determines what to store by identifying evidence in new observations that cannot be recovered from existing memory. The challenger analyzes the latest observation and generates question-answer pairs, termed challenges, to assess whether the observed information is captured in memory. The predictor attempts to answer each challenge using current memory, while the judge evaluates the predicted answers against the ground-truth answers provided by the challenger. 
Correctly answered challenges trigger reinforcement of retrieved memory nodes with the current observation timestamp, without duplicating stored evidence. Incorrectly answered challenges instead identify missing information for insertion or updates. 
We implement the challenger, predictor, and judge modules using a VLM-based agents, as illustrated in Fig.~\ref{fig:arch}.

Prior to memory initialization, we configure the challenger with $I = \{(\gamma_k, \epsilon_k)\}_{k=1}^{K}$, consisting of $K$ challenge categories $\gamma_k \in \mathcal{T}$, each accompanied by $E$ example questions $\epsilon_k \in \mathcal{T}^E$. 
These categories and examples can be generic or tailored to the target task.
For each new observation $O_t$, the challenger generates $N$ challenges in the form of question-answer pairs, given the observation, the current memory $M_{t-1}$, and the configuration $I$:
\begin{equation}
    (Q_t, A_t) = C(O_t, M_{t-1}, I),
\end{equation}
where $Q_t = (q_t^n)_{n=1}^{N}$ and $A_t = (a_t^n)_{n=1}^{N}$ denote the sequences of questions and their corresponding reference answers, respectively, with $q_t^n, a_t^n \in \mathcal{T}$. The predictor answers the challenge questions $Q_t$ using the current memory 
\begin{equation}
    \tilde{A}_t = P(Q_t, M_{t-1}),
\end{equation}
where $\tilde{A}_t = (\tilde{a}_t^n)_{n=1}^{N}$ denotes the sequence of predicted answers, with each $\tilde{a}_t^n \in \mathcal{T}$ corresponding to question $q_t^n$. 
Subsequently, the judge evaluates the predicted answers against the challenger's reference answers to produce a binary correctness signal:
\begin{equation}
    D_t = \Delta(\tilde{A}_t, A_t),
\end{equation}
where $D_t = (d_t^n)_{n=1}^{N}$, with $d_t^n \in \{0,1\}$ recording predictor's correctness to the $n$-th challenge. 
Finally, we collect all challenges and their correctness signals into evidence
\begin{equation}
    E_t = \left(Q_t, A_t, \tilde{A}_t, D_t, O_t \right).
\end{equation} 
The resulting evidence is passed to the polymorphic branch.


\subsection{Polymorphic Branch}
\label{sec:poly}

The polymorphic branch determines how to store and organize knowledge in memory. It comprises an updater that incorporates new evidence from each observation and a reorganizer that continuously restructures the memory by consolidating recurring patterns into higher-level routines and abstractions. Both are implemented as VLM-based agents, as illustrated in Fig.~\ref{fig:arch}. For each new observation $O_t$, the updater processes the evidence $E_t$ in predictive branch to update the memory:
\begin{equation}
    M_t = U(E_t, M_{t-1}).
\end{equation}
The updater incorporates missing information from incorrectly answered challenges ($d_t^n = 0$), while correctly answered challenges reinforce the retrieved nodes by appending the current observation timestamp. Thus, recurring evidence is preserved without creating redundant memory nodes.
In parallel, the reorganizer asynchronously restructures the existing memory:
\begin{equation}
    M_t' = R(M_t).
\end{equation}

Both processes modify the memory by inserting, modifying, merging, or removing nodes and relationships, resulting in updated node and edge sets $\mathcal{N}'$ and $\mathcal{E}'$, respectively. 
The updater derives these changes from challenge outcomes and existing memory, whereas the reorganizer operates solely by analyzing the existing memory.

\subsection{Memory Retrieval}
\label{sec:qa}

At question time $t$, the answerer receives the latest observation from the video $O_t$ and the memory $M_{t-1}$ constructed so far. Given a question $x \in \mathcal{T}$, the answerer retrieves relevant information from memory and combines it with the current observation to generate an answer:
\begin{equation}
    \hat{y} = A(x, M_{t-1}, O_t),
\end{equation} 
where $A$ denotes the VLM-based answerer and $\hat{y} \in \mathcal{T}$ the generated answer. The observation $O_t$ contains the same modalities as during memory construction but may span a different temporal window. All VLM-based agent modules access the memory through a shared set of retrieval operations. These include inspecting a compact outline of the memory hierarchy, reading nodes and their relationships, and searching for relevant nodes through lexical and semantic matching, optionally restricted to specific temporal intervals. The agents can iteratively retrieve and inspect more nodes based on the information gathered so far. While the predictor and answerer use these operations to retrieve evidence for question answering, the updater and reorganizer use them to analyze the existing memory and determine where new nodes and relationships should be inserted.

\begin{table*}[t]
\centering
\small
\setlength{\tabcolsep}{4pt}
\caption{\textbf{Results across all benchmarks.}
Our \ours achieves the best performance on every metric.
All metrics are in \% except MultiHop scores (0--10).
``Long'' metric averages EgoLifeQA and MM-Lifelong, ``Other'' averages the remainder.
For UCS, we also report LEDGER's test split~\cite{chaudhari2026lookbackunderstandingpersistence}.
$^{a}$ denotes GPT-5 judge (otherwise gpt-oss-120b). $^{*}$ denotes author-reported results (otherwise our runs).
}
\vspace{-2mm}
\label{tab:main}
\resizebox{\textwidth}{!}{%
\begin{tabular}{l cccc ccc ccc}
\toprule
 & \multicolumn{4}{c}{Episodic}
 & \multicolumn{3}{c}{Spatial}
 & \multicolumn{3}{c}{Averages} \\
\cmidrule(lr){2-5}\cmidrule(lr){6-8}\cmidrule(lr){9-11}
 & & & \multicolumn{2}{c}{MultiHop-EgoQA}
 & \multicolumn{2}{c}{UCS} & & & & \\
\cmidrule(lr){4-5}\cmidrule(lr){6-7}
Method & EgoLifeQA & MM-Lifelong & IoU@0.3 & Score
       & LEDGER's & All & Beyond3D & Long & Other & All \\
\midrule
\multicolumn{11}{l}{\textit{General MLLMs}} \\
Cambrian-S 7B~\cite{cambrian_s} & 32.2 & 11.5 & 6.2 & 1.24
              & 38.4 & 38.9 & 35.0 & 21.9 & 26.7 & 24.8 \\
VideoChat-Flash 7B~\cite{videochat} & 36.2 & 14.8 & 6.0 & 1.96
                   & 41.5 & 42.0 & 35.3 & 25.5 & 27.8 & 26.8 \\
InternVideo2.5 8B~\cite{internvideo25} & 35.4 & 15.3 & 7.3 & 2.82
                  & 42.1 & 41.7 & 36.5 & 25.3 & 28.5 & 27.2 \\
Nemotron3.5-120B~\cite{nemotron35} & 36.2 & 12.0 & 20.8 & 3.36
             & 41.6 & 44.6 & 35.9 & 24.1 & 33.8 & 29.9 \\
Gemma4-31B-IT~\cite{gemma4} & 35.8 & 10.5 & \underline{38.8} & \underline{3.76}
        & \underline{48.2} & 50.2 & 22.0 & 23.2 & 37.0 & 31.5 \\
Qwen3.8-27B~\cite{bai2025qwen3vltechnicalreport} & 36.6 & 20.0 & 32.0 & 3.64
            & 47.9 & \underline{50.4} & 52.3
            & 28.3 & \underline{44.9} & 38.3 \\
\midrule
\multicolumn{11}{l}{\textit{Memory Methods}} \\
GEB$^{*}$~\cite{ren2026GEB} & 72.0 & 36.8$^{a}$ & 25.9 & 3.11
          & -- & -- & -- & -- & -- & -- \\
MERIT~\cite{choi2026keep} & 70.8 & \underline{34.0} & 17.0 & 3.19
      & 41.0 & 42.8 & 52.4 & 52.4 & 37.4 & 43.4 \\
MAGIC-Video~\cite{li2026magic} & \underline{71.2} & \underline{34.0} & 15.1
            & 3.73 & 44.6 & 48.6 & \underline{55.8}
            & \underline{52.6} & 39.8 & \underline{44.9} \\
WorldMM~\cite{yeo2026worldmm} & 66.8 & 32.4 & 21.5 & 3.22
        & 40.7 & 44.2 & 47.4 & 49.6 & 37.7 & 42.5 \\
DirectMe~\cite{wang2026keep} & 30.0 & 2.2 & 7.2 & 3.08
         & 39.8$^*$ & \underline{50.5}$^*$/42.1 & 34.8 & 16.1 & 28.1 & 23.3 \\
LEDGER~\cite{chaudhari2026lookbackunderstandingpersistence} & 30.8 & 9.2 & 12.8 & 2.56
       & 46.3$^*$ & -- & 34.4 & -- & -- & -- \\
\midrule
\textit{\ours (ours)} & \textbf{74.5} & \textbf{47.8} & \textbf{42.1}
              & \textbf{4.52} & \textbf{50.8} & \textbf{53.6}
              & \textbf{62.5} & \textbf{61.1} & \textbf{52.7}
              & \textbf{56.0} \\
\bottomrule
\end{tabular}%
}
\end{table*}

\section{Experiments}

\begin{figure*}[t]
  \centering
  \begin{minipage}[t]{0.325\textwidth}
    \centering
    \includegraphics[width=\linewidth]{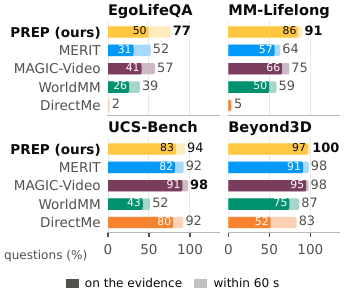}\\[-2pt]
  \end{minipage}\hfill
  \begin{minipage}[t]{0.325\textwidth}
    \centering
    \includegraphics[width=\linewidth]{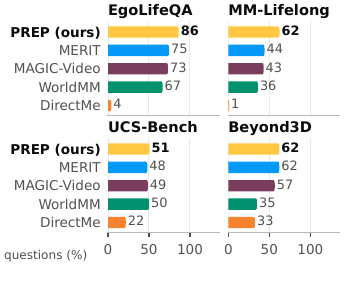}\\[-2pt]
  \end{minipage}\hfill
  \begin{minipage}[t]{0.325\textwidth}
    \centering
    \includegraphics[width=\linewidth]{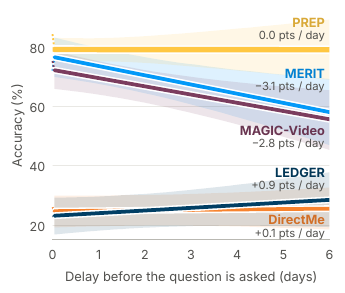}\\[-2pt]
  \end{minipage}
  \vspace{-2mm}
    \caption{\textbf{Memory hits and retention.}
\textbf{(Left, Middle)} Percentage of questions with correct supporting evidence retrieved from memory.
\textbf{(Left)} Evidence within 60\,s of ground truth, spanning less than 60\,s.
\textbf{(Middle)} VLM judgment given the question, ground-truth answer, and retrieved memory.
\textbf{(Right)} Answer accuracy for questions repeated after increasing delays as memory continues updating.}
  \vspace{-3mm}
  \label{fig:hits-retention}
\end{figure*}

\paragraph{Evaluation.}
We evaluate three episodic and two spatial benchmarks. For episodic memory, EgoLifeQA~\cite{yang2025egolife} contains 500 multiple-choice questions over 52\,h of week-long video from one of six participants. The Week split of MM-Lifelong~\cite{chen2026multimodallifelongunderstandingdataset} has 200 open-ended questions over same recordings, scored by a LLM judge. MultiHop-EgoQA~\cite{chen2025grounded} contains 1,080 open-ended questions with temporal grounding over 180-s clips. For spatial memory, UCS-Bench~\cite{wang2026keep} contains 8,114 questions over 532 videos. Beyond3D~\cite{ma2026beyond3d} contains 9,000 questions across 148 videos, evaluated by mean accuracy across question types.

\noindent\textbf{Implementation details.}
Our \ours continuously updates its memory online from successive 30\,s video clips sampled at 2\,FPS ($f_t=60$ frames per observation $O_t$). At answering time $t$, the model receives only the memory accumulated before $t$ and the last 60\,s of video sampled at 1\,FPS. 
Challenger categories are initialized from held-out question-answer pairs, or from 200--300 randomly sampled questions excluded from evaluation when no held-out set exists (UCS-Bench, Beyond3D).
All five modules of our \ours use Qwen3.8-27B (FP8). 
Baselines are evaluated using their released codebases and Qwen3.8-27B, unless otherwise stated. All experiments run on a single NVIDIA H200 GPU. 
We evaluate on each benchmark following the released protocol, and matching it for all our reproduced baselines.
We calculate 95\% bootstrap confidence (CI) intervals by resampling inputs of different methods to assess the statistical significance of performance differences. The Appendix provides additional details.

\subsection{Main Results}
\label{sec:main}

\noindent\textbf{Question Answering.}
Table~\ref{tab:main} evaluates three episodic and two spatial benchmarks.
Our \ours leads on all five, with non-overlapping 95\% bootstrap confidence intervals against the strongest baseline on four. On EgoLifeQA, our CI [71.1, 77.6] overlaps with MAGIC-Video [67.2, 75.0] and MERIT [66.6, 74.8].
Most methods excel on their target benchmarks but degrade substantially on others. MAGIC-Video is a notable exception, performing well across tasks but consistently behind \ours. Moreover, \ours is the only method that consistently outperforms the Qwen3.8-27B base model across all benchmarks (17.7pts on average).

\noindent\textbf{Memory Retrieval.}
We evaluate whether methods retrieve correct evidence from memory, independently of answer accuracy, using two complementary measures. Following~\cite{choi2026keep}, Fig.~\ref{fig:hits-retention} (a) counts a memory hit only when retrieved evidence spans less than 60\,s and falls within 60\,s of the annotated ground truth. This prevents broad generalizations spanning long periods from trivially matching the ground truth, but also penalizes valid abstractions spanning multiple observations. To account for these, Fig.~\ref{fig:hits-retention} (b) instead uses an LLM judge (Qwen3.5-VL) to assess whether the retrieved memory contains sufficient evidence to answer the question, given the question and ground-truth answer. Both measures show that \ours{} achieves the best retrieval overall, with particularly large gains on the longer EgoLifeQA and MM-Lifelong benchmarks.

\noindent\textbf{Memory Retention.}
Benchmarks evaluate questions at predefined timestamps. To assess retention, we repeat 154 EgoLifeQA questions at varying intervals after their original timestamps, while memory continues to update. Fig.~\ref{fig:hits-retention} shows that \ours maintains its performance over time, while competing methods progressively degrade.

\subsection{Efficiency}
\label{sec:efficiency}

\noindent\textbf{Real-Time Streaming.}
We define real-time streaming as continuously building memory from incoming video clips without falling behind the video stream. \ours processes consecutive 30\,s clips, discarding each after updating memory. On a single H200 GPU, the predictive and polymorphic stages average 8.9\,s and 14.4\,s, respectively (maxima 11.7\,s and 19.4\,s), totaling 23.3\,s per clip. The predictive stage always finishes before the next clip arrives, while 78\% of complete updates also finish within 30\,s. Otherwise, polymorphic updates continue asynchronously as the predictor processes the next clip, maintaining uninterrupted streaming. For computational efficiency during evaluation, we batch consecutive clips during memory updates, trading memory update latency for faster throughput. Processing speed remains stable as memory grows to 9,126 nodes over a week of EgoLife.

\noindent\textbf{Memory Storage.}
Table~\ref{tab:eff-summary} shows that our \ours storing only text requires just 0.93\,MB per hour. Baselines require 5--17\,MB/h for textual memory plus 1--2\,GB/h for retained video evidence. Thus, \ours uses three orders of magnitude less storage. Over a week of continuous streaming, this amounts to 0.15\,GB versus 169--334\,GB.

\noindent\textbf{Answering Cost.}
Table~\ref{tab:eff-summary} shows that \ours requires 14.0 GPU-seconds per question, matching MAGIC-Video and substantially faster than other baselines (25.8--79.1\,s). It also uses fewer model calls (2.7 vs.\ 3.2--3.9), except DirectMe, which uses a single call by design but processes all frames, requiring 79.1 GPU-seconds per question.

\noindent\textbf{Efficiency Trade-offs.}
Longer observation clips substantially reduce processing cost with little accuracy impact. On EgoLifeQA, increasing clip length from 5 to 30\,s reduces memory building from 38.7 to 8.9 GPU-hours, while accuracy remains comparable.
In contrast, limiting memory retrieval to a single search reduces accuracy by 5.7 and 17.8 points on EgoLifeQA and MM-Lifelong, respectively. Nevertheless, \ours remains the best-performing method when all methods are restricted to one search.

\begin{table}[t]
  \centering
  \small
  \footnotesize
  \setlength{\tabcolsep}{3pt}
    \caption{\textbf{Memory storage efficiency and answering cost.}
    }
    \vspace{-3mm}
  \label{tab:eff-summary}
  \resizebox{\columnwidth}{!}{%
  \begin{tabular}{@{}l cc c c c@{}}
    \toprule
    & \multicolumn{2}{c}{Memory [MB/h]} & & & Tokens \\
    \cmidrule(lr){2-3}
    Method & text & total & Calls & GPU-s & (k) \\
    \midrule
    MERIT~\cite{choi2026keep}              & 5.03  & 1985 & 3.21 & 25.8 & 11.1 \\
    MAGIC-Video~\cite{li2026magic}         & 13.17 & 1993 & 3.91 & \textbf{14.0} & 9.8 \\
    WorldMM~\cite{yeo2026worldmm}          & 9.06  & 1989 & 3.75 & 73.2 & 53.7 \\
    DirectMe~\cite{wang2026keep} & 17.05 & 1033 & \textbf{1.00} & 79.1 & \textbf{0.4} \\
    \midrule
    \ours{} (ours)     & \textbf{0.93} & \textbf{0.93} & 2.66 & \textbf{14.0} & 7.0 \\
    \bottomrule
  \end{tabular}}
\end{table}

\begin{figure}[t]
  \centering  
  \includegraphics[width=\columnwidth]{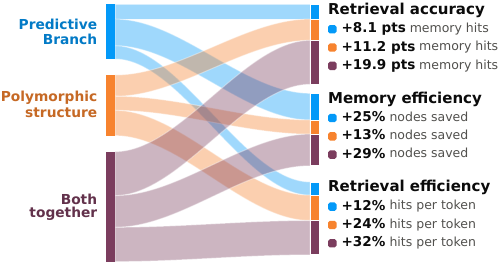}
  \vspace{-6mm}
    \caption{\textbf{Component Ablation.}
    Removing the predictive branch replaces challenger evidence with captions. Removing the polymorphic branch replaces hierarchical abstraction with a flat  list.}
  \vspace{-3mm}
  \label{fig:ablation}
\end{figure}

\subsection{Ablation Study} 
\label{sec:ablation}

\paragraph{Component Ablations.}
We ablate the predictive branch by removing the challenges, so the same writer records each window directly. We ablate the polymorphic branch by disabling restructuring, which leaves a flat list of memory nodes without hierarchy or abstractions. Fig.~\ref{fig:ablation} shows that removing these components reduces memory retrieval accuracy by 8.1 and 11.2 points, respectively, and by 19.9 points when both are removed. The predictive branch filters redundant observations, reducing stored nodes by 25\%, while the polymorphic branch consolidates recurring evidence into abstractions, reducing nodes by 13\%. This also improves building efficiency, as processing the $t$-th hour costs $T(t)=G+a\rho t$, where $G$ is the initial hourly cost, $\rho$ the node growth rate, and $a$ the cost per stored node. Furthermore, both components improve efficiency by reducing redundant evidence and simplifying retrieval.
Moreover, in a separate study, we ablate the challenger's initialization with categories derived from held-out question-answer pairs. Disabling this initialization reduces QA accuracy by only 0.1 points on EgoLife and 1.3 points on MM-Lifelong, demonstrating that performance is largely independent of category priors. More details in appendix.

\begin{figure}[t]
    \centering
    \includegraphics[width=\linewidth]{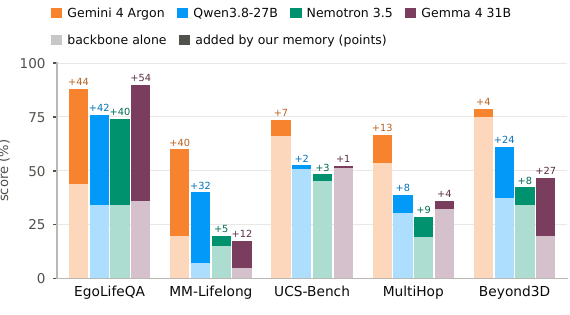}
    \vspace{-7mm}
    \caption{\textbf{Backbone Sensitivity.} Results use 10\% of each benchmark. Light bars show base-model performance. Darker segments show gains from each model as the \ours{} backbone.}
    \vspace{-4mm}
    \label{fig:backbones}
\end{figure}

\subsection{Behavior Study}
\label{sec:behavior}

\noindent\textbf{Backbone Sensitivity.} We evaluate \ours{} with four different backbones on 10\% of each benchmark (Fig.~\ref{fig:backbones}), comparing against the corresponding base models given uniformly sampled video frames. Our \ours{} framework consistently improves performance across all backbones, with the largest gains on long-duration benchmarks such as EgoLifeQA and MM-Lifelong. Gemini 4 Argon performs best overall, both with and without memory. Stronger backbones also produce better memories that transfer to weaker models. Qwen3.8 improves when using Gemini-generated memory instead of its own, from 52.6 to 58.0 on UCS-Bench and 38.9 to 67.6 on MultiHop-EgoQA. However, Gemini still performs best when querying its own memory.

\begin{figure*}[t]
  \centering
  \includegraphics[width=0.99\textwidth]{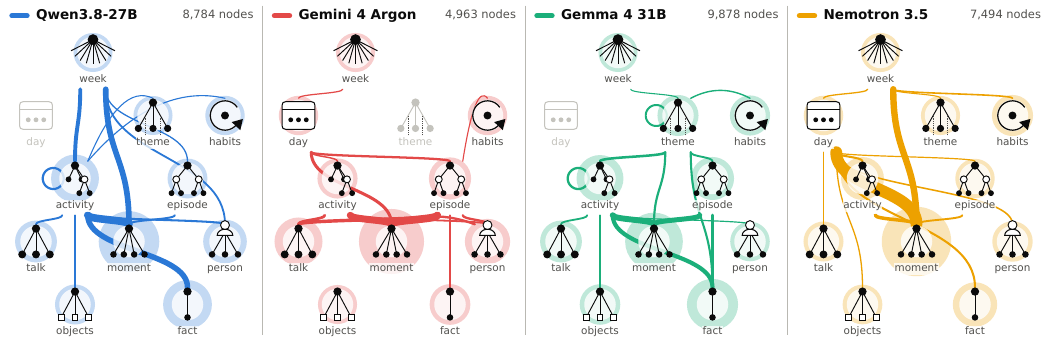}
  \vspace{-2mm}
    \caption{\textbf{Emergent memory topologies.} Memory hierarchies independently formed by \ours{} with different backbones on the same video, without an imposed structure. The figure highlights emergent memory structures rather than cross-comparing backbone topologies.}
  \vspace{-3mm}
  \label{fig:structures}
\end{figure*}

\noindent\textbf{Emergent memory topology.}
Although \ours{} imposes no fixed structure, all four backbones independently develop hierarchical memories on the same video (Fig.~\ref{fig:structures}). We identify meaningful semantic groupings (e.g., days, conversations, habits) by embedding memory nodes into latent space and verifying that these patterns disappear under random rewiring. Repeated Qwen3.8 runs recover consistent unique hierarchies. Further analysis is provided in the appendix.

\noindent\textbf{Emergent memory abstractions.}
As \ours{} progressively consolidates observations, abstracted nodes emerge capturing recurring patterns across clips. Fig.~\ref{fig:abstraction} illustrates a node summarizing a group's running joke spanning six days, which cannot be inferred from any individual observation.

\begin{figure}[t]
  \centering
  \fbox{\parbox{0.94\columnwidth}{\small
  \textbf{General node} \hfill \emph{6 days, 10 people}\\[3pt]
  \textbf{The ``talent show'' running joke.} Someone says ``show some talent'', and the group either
  brushes it off or performs on the spot. Shure is the one who usually performs.\\[3pt]
  \emph{Seen in:}\\
  \begin{tabular}{@{}l@{\hspace{6pt}}l@{}}
    DAY3, night & guitar and singing in the courtyard\\
    DAY4, dusk & guitar and singing in the courtyard\\
    DAY4, evening & guitar set in the living room\\
    DAY6, night & small concert in the second-floor living room\\
    ...
  \end{tabular}}}
  \vspace{-2mm}
  \caption{\textbf{Emergent memory abstraction.} A high-level node directly below the root, progressively consolidated to capture a group's recurring joke across days of observations.}
  \label{fig:abstraction}
\end{figure}

\begin{figure}[t]
  \centering
  \includegraphics[width=0.95\columnwidth]{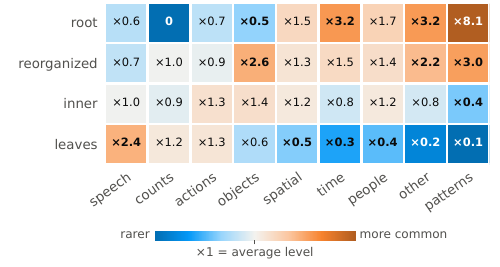}
\vspace{-2mm}
\caption{\textbf{Node prevalence across levels.} Each cell shows category prevalence relative to its four-level average ($\times 2$ means twice the average). Speech dominates leaves, objects reorganized nodes, and temporal facts and recurring patterns the root.}
\vspace{-5mm}
  \label{fig:depth}
\end{figure}

\section{Discussion}
\label{sec:discussion}
We never ask \ours to build a tree. Memory operations only constrain nodes to be inserted under another node or above the root, extended, renamed, moved, or merged. These operations permit a flat list, a timeline, or a deep taxonomy. Nevertheless, all backbones develop hierarchical memory structures (Fig.~\ref{fig:structures}). This emergent organization facilitates both memory reorganization and retrieval. By grouping related evidence into subtrees, it simplifies identifying nodes to merge and consolidate into higher-level abstractions, without searching every node (Fig.~\ref{fig:ablation}). Similarly, during retrieval, the agent can navigate relevant branches, while higher-level abstractions capture recurring patterns without requiring traversal of individual observations (Fig.~\ref{fig:abstraction}). This suggests that the agent naturally develops a structure that facilitates its future interactions with memory, without an explicitly prescribed hierarchy. Moreover, the same flexible organization could also support richer memory representations. While our experiments demonstrate the effectiveness of text-only memory, selectively storing frames, image crops, or learned visual features could further preserve information difficult to express in text.

\section{Limitations}
Our challenge-driven memory construction may overlook novel information not covered by generated questions. The text-only memory can lose fine-grained visual details, while repeated VLM-based memory updates and abstractions risk accumulating errors. Finally, despite matching the fastest answering time among compared methods, inference still takes 14 seconds per question on an H200 GPU, with substantial token consumption, limiting practical deployment.

\section{Conclusion}
We introduced \ours, a training-free framework for long-form egocentric video memory that jointly determines what to remember and how to organize it. Its predictive branch identifies information that cannot be recovered from existing memory, while its polymorphic branch organizes knowledge into an evolving hierarchy of events, facts, and higher-level abstractions without a predefined schema. Unlike existing memory methods that excel on specific tasks but fail to consistently outperform general-purpose VLMs, \ours achieves state-of-the-art performance across five episodic and spatial benchmarks, consistently surpassing the underlying VLM. It also supports real-time streaming with less than 1 MB of storage per video hour. Our experiments further demonstrate superior evidence retrieval and long-term retention, while ablations reveal that predictive updates and hierarchical abstractions jointly improve memory accuracy and efficiency. These findings suggest that effective egocentric memory depends not on storing more observations, but on selectively retaining information and organizing it at the appropriate level of abstraction. While demonstrated on egocentric videos, these principles are not specific to this setting and may extend to other long-form videos and streaming sensor modalities.

\section*{Acknowledgments}
This research was partially funded by the Ministry of Education and Science of Bulgaria (support for INSAIT, part of the Bulgarian
National Roadmap for Research Infrastructure). This work is supported by Google.

\clearpage

\section*{LLM Usage Statement}
In this work, we used LLMs (Claude Opus 5.5, Claude Fable 5.2, GPT6-Astra) to edit and refine author-produced text and improve readability. We also used AI tools to generate material for code and implementation, experiments, and figures and diagrams. The authors reviewed, edited and approved this material. We did not use LLMs or other AI tools for synthetic data generation, dataset creation or annotation, or to develop research ideas, hypotheses, claims or conclusions. All AI-assisted work was reviewed by the authors. We take full responsibility for the final content of this work, including all text, code, experimental results, and visual material produced with AI assistance.

{
    \small
    \bibliographystyle{ieeenat_fullname}
    \bibliography{main}
}

\clearpage
\setcounter{page}{1}
\maketitlesupplementary
\setcounter{section}{0}
\renewcommand{\thesection}{\Alph{section}}
\setcounter{figure}{0}
\renewcommand{\thefigure}{S\arabic{figure}}
\setcounter{table}{0}
\renewcommand{\thetable}{S\arabic{table}}

\etocdepthtag.toc{supp}
\etocsettagdepth{main}{none}
\etocsettagdepth{supp}{subsubsection}
\providecommand{\supptocline}[5]{}
\renewcommand{\supptocline}[5]{%
  \par\begingroup
  \normalfont\normalsize
  \parindent=0pt\parskip=0pt
  \leftskip=\dimexpr#1+#2\relax
  \rightskip=2.5em\parfillskip=-2.5em
  \leavevmode\llap{\makebox[#2][l]{#3}}%
  {#4}\nobreak\hskip0.6em
  \leaders\hbox to 0.5em{\hss.\hss}\hfill
  \nobreak\hbox to 2.5em{\hfil #5}\par
  \endgroup}
\etocsetstyle{section}{}{}{\smallskip\supptocline{0pt}{2.1em}{\textbf{\etocnumber}}{\textbf{\etocname}}{\textbf{\etocpage}}}{}
\etocsetstyle{subsection}{}{}{\supptocline{1.0em}{3.0em}{\etocnumber}{\etocname}{\etocpage}}{}
\etocsetstyle{subsubsection}{}{}{\supptocline{2.0em}{4.0em}{\etocnumber}{\etocname}{\etocpage}}{}
\tableofcontents

\section{Implementation Details}
\label{sec:supp-impl}

\subsection{System and Tools}
\label{sec:supp-system}

\noindent\textbf{Forming the observation.}
The stream is cut into clips of 30 seconds. The observation of a clip is $O_t = (V_t, S_t, C_t)$: 60 frames (2 FPS, 512 pixels), the transcript of the speech, and a caption. Beyond3D is released at 1 FPS, so a clip there has 30 distinct frames. Frames are decoded from the video when the clip arrives. Nothing is prepared ahead of the stream.

\emph{Transcript.} EgoLife releases transcripts with speaker names, and we use them. On the other benchmarks we transcribe the audio with Whisper large-v3-turbo. The audio is resampled to 16\,kHz mono and transcribed in windows of 30 seconds with 5 seconds of overlap. Each window is transcribed on its own, as soon as its audio has arrived, so transcription never needs future audio. Silent windows are skipped, the language is detected per window, and repeated text in the overlap is removed. Beyond3D is silent, so $S_t$ is empty there.

\emph{Caption.} EgoLife releases dense captions, and we use them. On the other benchmarks the Challenger writes the caption. It already looks at the 60 frames of the clip to write its challenges, and it returns the caption in the same call. Captioning therefore adds no model call and no second pass over the video, and its cost is included in every build cost we report (\cref{sec:supp-cost}).

\noindent\textbf{Models.}
All five modules and the Answerer use Qwen3.8-27B (FP8). Whisper large-v3-turbo transcribes speech. Qwen3-Embedding-4B (first 512 dimensions) embeds the memory for the Answerer's search by meaning. No other model is used to build or query the memory, and no model is trained or fine-tuned.

\noindent\textbf{Memory.}
The memory is a rooted structure of text nodes. A node has a title, a body, a time span, parents and children. The prompts require a rooted structure and generality towards the root, but specify no node types, branching or depth, thus allowing for any structure of the sort (tree, linked list, taxonomy, timeline). Within these constraints, the tree's organization emerges through updates and reorganization rather than following a predefined hierarchy. The Updater can \emph{add} a node and \emph{append} to a body. The Reorganizer can also \emph{retitle}, \emph{move} and \emph{merge} nodes, and can add a more general node above the root. A merge removes the duplicate node. Every operation is written to a log, so the memory at any past time can be restored exactly. The memory also keeps the transcript and caption of each clip as text, verbatim, under its time range (the \emph{record}). It is part of the stored text and of the size we report. The memory holds no frames.

\noindent\textbf{Memory outline.}
The outline lists the nodes of the tree level by level from the root (identifier, title, time span, and number of children, no contents) up to a fixed character budget. It is obtained by a breadth-first traversal of the current memory, emitting these fields until the budget is reached. It exposes the organization of the tree; node bodies are supplied separately through matched nodes or read operations.

\noindent\textbf{Number of challenges.}
The Challenger asks three challenges per clip. Each has a question, a reference answer, the evidence for it, and one of the question categories (\cref{sec:supp-cats}).

\noindent\textbf{What each module reads and returns.}
\Cref{tab:supp-modules} lists the settings. Every call ends with a structured reply that is validated. If validation fails, the module gets the error and one more turn.
\begin{table}[!htb]
\centering
\small
\setlength{\tabcolsep}{4pt}
\caption{\textbf{Settings per module.} Turns are the maximum number of model turns in one call. Thinking is on only for the Answerer. No call has an output-token limit.}
\label{tab:supp-modules}
\begin{tabular}{lccl}
\toprule
Module & Turns & Temp. & Input \\
\midrule
Challenger & 5 & 0.7 & frames, text, memory \\
Predictor & 2 & 0.0 & memory \\
Judge & 2 & 0.0 & text \\
Updater (check) & 8 & 0.0 & memory, tools \\
Updater (write) & 4 & 0.3 & frames, text, memory \\
Reorganizer & 8 & 0.5 & memory, tools \\
Answerer & 8 & 0.6 & frames, memory, tools \\
\bottomrule
\end{tabular}
\end{table}

\begin{itemize}
\item \textbf{Challenger}: reads the 60 frames, the clip's text, and a view of the memory (outline from the root, nodes matching the clip, 8{,}000 characters), with the categories and up to five example questions each. Returns three challenges and, when needed, the caption.
\item \textbf{Predictor}: reads the three questions and the view of $M_{t-1}$. It sees no frame, no text of the clip and no reference answer. The outline and the search for matching nodes are run once for it. Returns one prediction per question with the nodes it rests on.
\item \textbf{Judge}: reads questions, reference answers and predictions as text. Returns a correctness verdict per challenge. An answer that is only partly right counts as failed. The challenges with their verdicts form the evidence $E_t$.
\item \textbf{Updater}: first checks, with read and search tools, whether the memory already contains the answer of a failed challenge. If it does, the nodes that store it are flagged for the Reorganizer: the fact was stored but not found. Otherwise the Updater reads the failed challenges, the clip texts and up to 32 frames, and returns add and append operations.
\item \textbf{Reorganizer}: runs in the background on the live memory. It first revisits the nodes flagged by failed questions whose answer the memory already held, then scans the structure level by level and writes the generalizations that the specifics support.
\item \textbf{Answerer}: reads the question, the frames the benchmark gives with it, and the view of the memory at the question time. It can read nodes, search by words or by meaning, list what happened between two times, and read the record.
\end{itemize}

\noindent\textbf{Coordination.}
The order of the calls for one clip is fixed: Challenger, Predictor, Judge, then Updater. The Reorganizer is the only module that runs independently of the clips. Updates are applied to the memory in stream order, one at a time, and each application is atomic. When a prediction is correct, the nodes it cited receive a dated ``seen again'' line and no new text is written. Because the Updater only adds and appends, and only the Reorganizer moves nodes, the two can work on the same memory without conflict. Node identifiers are never removed. A merge leaves the old identifier as an alias.

\noindent\textbf{Deployment.}
One job uses one NVIDIA H200. The model is served with vLLM~0.30.0 with a context of 65{,}536 tokens and up to 70 images per request (80 in the keyframe ablation \cref{sec:supp-visual}). At answer time the embedding model runs on the same GPU. A long stream is processed by a chain of jobs. Each job resumes from the last committed state of the operation log, so an interruption loses no work.

\begin{table*}[!t]
\centering
\small
\setlength{\tabcolsep}{4pt}
\begin{tabular}{@{}lp{14.2cm}@{}}
\hline
Tool & What it does \\
\hline
\multicolumn{2}{@{}l}{\emph{Reading the memory}} \\
\texttt{read} & Returns one node: its title, time span, body and children. \\
\texttt{search} & Returns the nodes whose title or body share words with the query, best match first. It can be restricted to a time range, or rank by the latest or earliest occurrence. \\
\texttt{timeline}$^\ast$ & Returns what the memory holds between two times, in time order. When too many nodes overlap, only the coarsest are shown. \\
\texttt{record}$^\ast$ & Returns the stream's saved caption and transcript, as the memory received it. It can be queried by time range or by words. \\
\texttt{similar}$^\ast$ & Searches by meaning instead of by words (with an embedding model), over nodes and recorded moments. It can be restricted to a time range. \\
\hline
\multicolumn{2}{@{}l}{\emph{Writing the memory}} \\
\texttt{add} & Adds a new node under a parent, with a title, a body and a time span. Without a parent, the new node goes above the current root, which becomes its child. \\
\texttt{append} & Adds text to the end of a node's body. Bodies only grow, so a change of state is written together with its time. A given span widens the node's span. \\
\texttt{retitle} & Gives a node a new title, and optionally a new span. \\
\texttt{move} & Moves a node, with everything below it, under another parent. \\
\texttt{merge} & Folds a duplicate into its twin. The duplicate's children and text move over and its id becomes an alias, so nothing is lost. \\
\hline
\end{tabular}
\caption{Tools. $^\ast$Answerer only. The roles that build the memory never read the stream again through
these tools. The Updater's check calls \texttt{read} and \texttt{search}. Its writing step submits \texttt{add} and \texttt{append} operations.
The memory refuses impossible edits, e.g., moving a node under its own descendant. No text is ever lost.}
\label{tab:supp-tools}
\end{table*}

\noindent\textbf{Tools.}
\label{sec:supp-tools}

We provide details on the tools the different models can call in Table~\ref{tab:supp-tools}.
\cref{sec:qa} lists the retrieval operations of the whole system. Each module uses a subset of them. The Challenger and the Predictor call no tools. They receive a view of the memory: its outline and the nodes found by one search, run once for them. The Updater's check and the Reorganizer read nodes and search by words over several turns. Only the Answerer searches by meaning, lists what happened between two times, and reads the record.

\Cref{sec:supp-update-example} walks through one complete memory update.

\subsection{Agentic Instructions}
\label{sec:supp-prompts}

All benchmarks use the same agentic instructions for each module. Only
benchmark-specific context and settings, such as the required answer format, are adapted. We reproduce
the shared instructions below. In the prompts, a \emph{window} is a clip
and a \emph{probe} is a challenge. The Challenger prompt is followed
by the category list (\cref{tab:supp-catlist}), and the Answerer
prompt by the answer format of the benchmark. The organization of
the memory is stated in one place, the memory rules, shared by the
Challenger, the Updater and the Reorganizer. The rules ask for
generality towards the root but fix no node types, branching or depth.
The tree's organization emerges during streaming.

\begin{promptbox}{Memory rules (shared)}

\textbf{How the memory is organized.}

The memory is a rooted structure of text nodes, more general towards
the root: the root holds what is true of the whole stream, each node
is more specific than its parent, single episodes are children on
the level farthest from the root. No fixed schema; a body states
abstractly what holds for all below it. Readers walk down from the
root: titles must show at a glance where things live.

\begin{promptitems}
  \item
  A fact goes at the highest level that covers it. A one-off moment
  under the node it is an instance of.

  \item
  Never remove occurrences: a state change or a pattern seen again
  is a new dated line (``keys: moved to the drawer at DAY2 14:10'').
  Two nodes that are one thing merge.

  \item
  Generalize: several nodes sharing a pattern get a pattern node,
  under the frame it belongs to, with the instances moved under it.
  A node above the root must be a more general frame of everything,
  never a day, task, object, moment.

  \item
  An episode is one fact at its moment, span in stream seconds.
  Each occurrence of an event is its own fact. What deserves one is
  what the questions asked of this memory need. Spans above are
  derived, never written.

  \item
  Write in english. Keep names, times and quoted speech exact.
\end{promptitems}

\end{promptbox}

\begin{promptbox}{Challenger}

Probe a memory being built from a first-person video stream.
Given one window (time-labelled frames, text) and the current
memory, ask about this window what the questions this memory will
be asked later (the category examples) will need from it, the way
the examples ask, exposing what the memory lacks, has wrong or
stores at the wrong level.

\begin{promptitems}
  \item
  \texttt{now}: answerable from this window alone.

  \item
  \texttt{earlier}: links this window to the memory shown; quote
  the node and give its id in \texttt{cites}. Nothing related in
  the memory shown: ask \texttt{now}; never invent a link.
\end{promptitems}

One question, one checkable answer: no comparison, no ``and'',
no ``is this the same X''. Name people only as the text does;
unnamed: ask what they do, hold or where they are. Give answer,
evidence and one category per question. When the window lists
categories, ask one question per listed category; otherwise cover
every category over the stream.

A requested caption describes the window as fully as those
questions need: people (names when known), actions, objects,
places, times.

\end{promptbox}

\begin{promptbox}{Predictor}

Predict the answers to questions about a moment of a first-person
video stream not yet in the memory, from what the memory knows
(habits, layout, ongoing activity, roles). State what the memory
shown (root, outline of the memory, matching nodes) makes likely,
as specifically as it supports, with the ids of the nodes it
rests on in \texttt{cites}.

No ground in the memory: ``unknown'', no cites.
Never general knowledge alone.

Call \texttt{submit\_answers}, one prediction per question, in order.

\end{promptbox}

\begin{promptbox}{Judge}

Grade answers given from memory alone against the reference answer
and evidence.

\begin{promptitems}
  \item
  \texttt{correct}: the same fact (wording may differ; an option
  letter equals its text);

  \item
  \texttt{partial}: incomplete, hedged or partly wrong;

  \item
  \texttt{wrong}: otherwise, including ``the memory does not hold
  this''.
\end{promptitems}

Call \texttt{submit\_verdicts}: one verdict and one-line reason per
question, in order.

Predictions were made from the memory before the moment was seen:
right by generalization is \texttt{correct} (memory: at the desk
every morning; reference: at the desk); ``unknown'' is
\texttt{wrong}.

\end{promptbox}

\begin{promptbox}{Updater}

\textbf{Check of the memory.}

For questions the memory could not predict, check whether it
already holds the answer. Search it by the reference answer's
names, places, objects and words; read the candidates whole.

\texttt{nodes}: ids of the nodes that state the reference answer
(the same fact, as a habit, layout, role or earlier instance that
still holds). \texttt{[]} when none.

Call \texttt{submit\_diagnoses}, one per question, in order.

\medskip
\textbf{Writing.}

Maintain the memory of a first-person video stream from the
evidence below (text, captions, frames). The questions the memory
failed show what it is asked and lacks; their answers are examples.
Write what the evidence shows that such questions would need,
as your own description of the scene, not a copy of the answers,
and nothing else: \texttt{add} nodes (\texttt{tmp} name, usable as
parent by later adds) and \texttt{append} to nodes by the ids shown.

\begin{promptitems}
  \item
  A new fact is an episode, a child on the level farthest from the
  root: \texttt{span} in stream seconds, the names the text gives,
  exact words when the words are the fact.

  \item
  Highest level that covers it: the root for what holds for the
  whole stream; a moment under the node it is an instance of,
  or the root when none fits.

  \item
  What continues the nodes written just before (shown with paths)
  goes under them: an activity, conversation or visit stays one
  node whose description and span grow (an append gives its span).
\end{promptitems}

Predicted facts are already recorded: do not re-describe them.
The questions below are the surprises; write what they show.

\end{promptbox}

\begin{promptbox}{Reorganizer}

Keep the memory of a first-person video stream general. Scan it
node after node, at every level, for generality the specifics
reveal that is not yet written; generality grows towards the root.

\begin{promptitems}
  \item
  The smallest generalization that cumulates specifics: a pattern
  over instances that actually share it (search for the others
  first), also across the whole stream.

  \item
  Title: the generalization as one sentence; body: its scope and
  its instances with times, so that it predicts the next occurrence.

  \item
  Record it where it holds: a pattern node over its instances
  (add, move them under it), the node whose frame it holds for,
  or the root. Always try to add to the root what has become general.

  \item
  Specifics answer most questions: never delete, rewrite or
  dilute them.

  \item
  Failed questions whose evidence the memory held were not found:
  fix what kept them from being found.

  \item
  On the way: generalize a title too narrow for its node,
  move misplaced nodes, merge duplicates, re-articulate a level
  whose titles no longer show where things live (the outline
  gives the width of every level).
\end{promptitems}

Use the tools only when there is something to do, then call
\texttt{done} with a one-line summary.

\end{promptbox}

\begin{promptbox}{Answerer}

Answer questions using only the memory of a first-person video
stream (shown: root, outline of the memory with level widths,
matching nodes with paths; tools: \texttt{read}, \texttt{search}).
Walk down from the outline when the subject is visible in it;
search otherwise.

Children are a node's specifics, its parent the activity, place
or pattern it belongs to, its span its time: pattern nodes answer
habits, routines and relations; an activity node the activity
and its time; its parts single moments.

Read candidates whole (search gives excerpts) until the question
is answered in full, every moment it rests on, and no further.
Times in nodes are exact. No general knowledge: when the memory
lacks the answer, say so with the closest thing it holds.
Follow the answer format given.

Call \texttt{submit\_answers}, one answer per question, in order.

\end{promptbox}

\subsection{Streaming and Batching}
\label{sec:supp-streaming}

\noindent\textbf{Causality.}
A call for clip $t$ reads only $O_t$ and the memory built from earlier clips. No call reads a later clip. A question asked at time $t$ is answered from the memory committed up to the clip that contains $t$. This holds in every experiment of the paper.

\noindent\textbf{Live sreaming.}
On a live stream each clip is processed when it ends. The Reorganizer runs in the background and never delays an update. Measured latency is reported in \cref{sec:supp-live-results}.

\noindent\textbf{Evaluation batching.}
The benchmarks are recorded videos, and processing 52 hours in real time would take 52 hours. A recorded video has all its clips available, so the GPU can serve many calls at once via clip batching and parallelization. We use this to build the same memory faster. It changes when calls are scheduled, not what they read.

\noindent\textbf{Concurrency.}
One stream is one asynchronous program, and all calls go to one model server, which batches the requests it holds. The following run concurrently:
\begin{itemize}
\item \emph{Updates.} Consecutive clips are grouped into batches of at most six, in stream order. The Updater makes one call per batch, and up to 12 batches are written at once. A batch closes early at a clip that holds a benchmark question, so the question is answered from a memory that is complete up to its clip. A batch with no failed challenge needs no call.
\item \emph{Reorganizer.} One call at a time, back to back, on the live memory.
\end{itemize}
One step is sequential: the results of the Updater are applied to the memory in stream order. This step makes no model call. A clip is not challenged while it is more than 38 clips ahead of the last applied update, so the Predictor always works from a recent memory.

Batching groups of clips that have already been seen, every call still reads only the past, updates are still applied in stream order, and each question is still answered from the state committed at its clip. A recorded video processed this way follows the same causal procedure as a live stream. 
On the other hand, in live streaming mode, there is no batching. There, clips are processed for memory updates as they appear.

\subsection{Benchmarks}
\label{sec:supp-bench}

\noindent\textbf{EgoLifeQA.}
500 multiple-choice questions (four options) on the week recorded by one wearer of EgoLife: 7 days, 51.95 hours. Each question has a time and is answered from the video before it. Five types: EntityLog (125), EventRecall (126), HabitInsight (61), RelationMap (125) and TaskMaster (63). Metric: accuracy.

\noindent\textbf{MM-Lifelong.}
The Week split of MM-Lifelong~\cite{chen2026multimodallifelongunderstandingdataset} has 200 open-ended questions over the same recordings, all asked at the end of the week. \ours{} answers them from the same memory as EgoLifeQA: one memory serves both benchmarks. We use the official scoring script unchanged. A judge model reads the question, the reference and the answer and gives a score from 0 to 5; scores 4 and 5 count as correct, 3 as half. The benchmark uses GPT-5 as the judge. We use the open model gpt-oss-120b, so that every number can be reproduced without a paid service, and we score all methods with it. We repeated the scoring with a second judge, Qwen3.8-Flash-Next. Judge agreement is reported in \cref{sec:supp-ci}.

\noindent\textbf{MultiHop-EgoQA.}
1{,}080 questions on 360 clips of 180 seconds from 258 Ego4D videos (18 hours). A reply gives an answer and the time intervals that support it. 724 questions have a reference answer. Grounding metrics, from the official script: IoU@0.3 (share of questions whose intervals overlap the reference by more than 0.3), mIoU, mIoP (precision) and mIoG (recall). The answer score is the official judge prompt with gpt-oss-120b.

\noindent\textbf{UCS-Bench.}
Multiple-choice, counting and yes/no questions about where things are and how the wearer moved, on videos from 10 seconds to 98 minutes. Each question has a time. Of the 532 released videos, 17 (293 questions) are set aside for the question categories and never scored. We report accuracy on the other 515 videos (7{,}821 questions, 167 hours), and on the subset of 204 videos (2{,}666 questions) used by LEDGER. Four categories: object recall and quantity, position and orientation, trajectory and movement, proximity and reachability.

\noindent\textbf{Beyond3D.}
9{,}000 multiple-choice questions on 148 kitchen videos of HD-EPIC (1 FPS, silent). Each question comes with the frame at its time; two types mark the object with a red point. Eight question types; the official score is the mean of the eight accuracies. 200 questions are set aside for the question categories, and the other 8{,}800 are scored in \cref{tab:spatial}. The averages of the main table include all 9{,}000 and differ by at most 0.2 points.

\noindent\textbf{Frames given with the question.}
Every benchmark defines the frames a model receives with a question, and \ours{} follows it. By default it receives the last 60 seconds at 1 FPS (EgoLifeQA, MM-Lifelong, and Beyond3D with the query frame). Where the released protocol fixes the frames, it receives those: the clip at one frame per three seconds (MultiHop-EgoQA) and the last 64 seconds at 1 FPS (UCS-Bench).

\begin{table*}[!t]
\centering
\footnotesize
\setlength{\tabcolsep}{4pt}
\caption{\textbf{Question categories per benchmark}, with their weight (out of 200 sampled questions).}
\label{tab:supp-catlist}
\begin{tabular}{@{}p{0.23\textwidth}p{0.23\textwidth}p{0.25\textwidth}p{0.22\textwidth}@{}}
\toprule
\textbf{EgoLife week} (9) & \textbf{MultiHop-EgoQA} (14) & \textbf{UCS-Bench} (14) & \textbf{Beyond3D} (8) \\
\midrule
Speech and conversation content (43)\newline
People and their actions (41)\newline
Objects and their states (28)\newline
Inference and reasoning (25)\newline
My own actions and activities (23)\newline
Temporal sequence and order (18)\newline
Spatial location and setting (11)\newline
Atmosphere and environment (10)\newline
Counts and quantities (1)
&
Temporal timing (72)\newline
Object manipulation actions (59)\newline
Placement and location (18)\newline
Opening and closing (13)\newline
Action sequences and order (7)\newline
Object identification by action (6)\newline
Tool and device operation (4)\newline
Repetitive actions (4)\newline
Counts and frequency (3)\newline
Cooking and food preparation (3)\newline
Carrying and transporting (3)\newline
Cleaning and maintenance (3)\newline
Other people involvement (3)\newline
Body part involvement (2)
&
Current spatial relations to self (35)\newline
Past action counts and frequencies (26)\newline
Reachability and accessibility (24)\newline
Past action details and object placement (24)\newline
Navigation and route planning (18)\newline
Scene object counts (17)\newline
Scene state and description (14)\newline
Trajectory and movement paths (10)\newline
Proximity comparisons (9)\newline
Object location (6)\newline
Plans, intentions, and what-to-do (6)\newline
Past reachability and past events (5)\newline
Object-to-object spatial relations (5)\newline
Routine and habitual patterns (1)
&
Visibility in current frame (45)\newline
Time object last stopped being moved (30)\newline
Distance from camera (24)\newline
Time object last visible (23)\newline
Distance between two objects (22)\newline
Relative position to marked object (21)\newline
Direction from viewpoint (18)\newline
Nearest fixture or counter area (17)
\\
\bottomrule
\end{tabular}
\end{table*}

\subsection{Challenger's Question Categories}
\label{sec:supp-cats}

\noindent\textbf{Procedure.}
The categories tell the Challenger which kinds of questions to ask. They are produced once per benchmark, before the stream, by one call of the backbone. The call reads 200 sampled questions as text, without video, and groups them into 8 to 15 categories. Each category has a name, a one-sentence description and a weight, the number of sampled questions it covers. Challenges are spread over the categories in proportion to the weights.

\noindent\textbf{No category leaking.}
The Challenger is shown up to five example questions per category. These examples never come from scored data: the EgoIT training questions for the EgoLife week (EgoLifeQA and MM-Lifelong), the Ego4D training split for MultiHop-EgoQA, the 293 and 200 held-out questions for UCS-Bench and Beyond3D. The Challenger never sees an answer to a benchmark question, and its challenges are about the clip in front of it.

\noindent\textbf{Category list.}
\Cref{tab:supp-catlist} prints all categories. They are kinds of information (speech, actions, object states, positions, counts, order), generic or tailored to the benchmark, and hold no answers. \Cref{sec:supp-ablation} shows that accuracy on the EgoLife week does not depend on them.

\begin{table*}[!t]
\centering
\footnotesize
\setlength{\tabcolsep}{4pt}
\caption{\textbf{How each method builds its memory from a long video and answers from it.} Size is per hour of EgoLife video.}
\label{tab:supp-methods}
\begin{tabular}{@{}p{0.09\textwidth}p{0.32\textwidth}p{0.2\textwidth}p{0.31\textwidth}@{}}
\toprule
Method & Building the memory & What is stored & Answering \\
\midrule
MERIT & Each 30-s clip gets a dense caption and four text keys (event, dialogue, object, summary), which are embedded. & Keys and embeddings (5.0\,MB/h) and the raw video. & An agent rewrites the query for up to 5 rounds, retrieves the top-5 clips and their neighbours, and answers from their frames. \\
MAGIC-Video & Captions at four time scales, speech, entity triples and clip embeddings form a graph; chains follow each entity and each multi-day event. & Graph and chains (13.2\,MB/h) and the raw video. & Up to 5 rounds of search over the graph, then an answer call with up to 64 retrieved frames. \\
WorldMM & 10-s captions are merged into multi-scale captions and graphs (episodic), consolidated triples (semantic) and clip embeddings (visual). & Three stores (9.1\,MB/h) and the raw video. & A controller chooses a store to search in each of up to 5 rounds; visual search returns up to 64 frames. \\
DirectMe & Depth, camera pose, object detection and segmentation on every frame at 1 FPS build a 3D graph of objects. & Object graph (17.05\,MB/h) and its keyframes at 1 FPS. & Retrieves the 32 closest objects, ranks their keyframes, and answers in one call from 32 keyframes. \\
LEDGER & Detection, camera poses and tracking lift objects to 3D. Places, relations and events are written as text. & 3D object tracks, camera poses and text (18.0\,MB/h), and the raw video. & One call reads the memory as text at the question time. With GPT-5.4, it also sees 48 frames sampled evenly from the video. \\
GEB & Tracking and re-identification build a biography per entity, linked in a graph of episodes. & Not reported. & Retrieval over the graph. \\
\midrule
\ours{} & For each 30-s clip: challenge, predict, judge. The failed challenges drive the update; the Reorganizer generalizes. & Text-based memory structure (0.93\,MB/h). No video. & One agent call that reads and searches the memory, with the frames the benchmark gives. \\
\bottomrule
\end{tabular}
\end{table*}

\subsection{Baselines}
\label{sec:supp-baselines}

\noindent\textbf{Baseline choices.}
We compare against the strongest released memory methods for long egocentric video, and against the methods proposed with our benchmarks. MERIT, MAGIC-Video and WorldMM are the best published methods with code on EgoLifeQA and MM-Lifelong. DirectMe is the method proposed with UCS-Bench, and LEDGER is a second 3D object memory published on it. GEB has the highest published EgoLifeQA score. However, its code is not released, so we quote the authors' numbers. The general models (Cambrian-S, VideoChat-Flash, InternVideo2.5, Nemotron, Gemma and our backbone Qwen3.8-27B) answer from the frames the benchmark gives, without memory. The row of our backbone shows what the memory adds.

\Cref{tab:supp-methods} summarizes the methods in the terms of our Sec.~4. All of them store a fixed structure chosen by their authors: an index of keys, a graph, three stores, or a 3D object map. All but LEDGER keep the raw video and retrieve frames from it at answer time. \ours{} is the only method that builds a text memory causally clip by clip, lets its structure emerge, and keeps no video.

\noindent\textbf{Matching baselines.}
The baselines never receive less than \ours{}, and on several points they receive more.
\begin{itemize}
\item \textit{Same video and same text.} All methods process the same recordings. On EgoLife, all methods use the same released transcripts and dense captions.
\item \textit{Same answering model.} The answers of the memory methods come from the same backbone as ours, with thinking on for all, unless the main table marks otherwise. Differences in score therefore come from the memory.
\item \textit{Same information at question time.} Every memory is cut at the time of the question, for all methods: no method reads video after the question. On Beyond3D, every method that reads images also receives the query frame.
\item \textit{The baselines see more video than ours.} MERIT, MAGIC-Video, WorldMM and DirectMe keep the whole video and read frames of any past moment when they answer, up to 64 frames per call. \ours{} keeps no video. At answer time it has its text memory and only the frames that the benchmark gives with the question.
\item \textit{Released code, unchanged.} Each baseline runs from its released code at a fixed commit, with its own prompts, retrieval settings and number of rounds. We change only the answer instruction, so that every method replies in the format the benchmark scores. As stated in Sec.~4 of the main text, every method follows the benchmark protocol, including its question times. DirectMe's released evaluation builds its graph from the whole video, so we feed it the video in stream order, and it answers each question only from the frames before it (\cref{tab:supp-cost}). Its 79.1 GPU-seconds per question in \Cref{tab:eff-summary} of the main text include this processing of all frames, which we did not time separately.
\item \textit{Same scoring.} All methods are scored by the same official scripts and the same judge.
\item \textit{No tuning on scored data.} The example questions shown to our Challenger come from data that is never scored (\cref{sec:supp-cats}), and a memory of the EgoLife week built without categories is as accurate (\cref{sec:supp-ablation}).
\end{itemize}

\subsection{Evaluation and Ablation Protocols}
\label{sec:supp-eval-protocols}

\noindent\textbf{Confidence intervals.}
\label{sec:supp-ci-impl}
Intervals are 95\% bootstrap intervals from 2{,}000 resamples (2.5th and 97.5th percentiles). The resampled unit is the question on EgoLifeQA and MM-Lifelong, and the video on MultiHop-EgoQA, UCS-Bench and Beyond3D, because questions on one video are not independent. A missing answer counts as wrong. All methods use the same resamples, so a difference between two methods is tested on paired draws. The scores of \ours{} on the EgoLife week are the mean of five answering passes. Results and intervals are reported in \cref{sec:supp-ci,fig:supp-ci}.

\noindent\textbf{Predictive branch.}
\label{sec:supp-predictive-impl}
The Predictor answers the three questions of a clip in one call. All three are answered from the same memory $M_{t-1}$, the state before the clip. On recorded videos, $M_{t-1}$ is the memory committed when the clip is challenged. On a live stream it is the memory after clip $t-1$. The call has no access to the clip, to the reference answers or to tools, and the memory is not updated between the three questions. A correct prediction therefore shows knowledge that the memory held before the clip, and nothing learned from one challenge can leak into another. Challenge statistics and update retention are reported in \cref{sec:supp-predictive}.

\noindent\textbf{Experiment-specific protocols.}
The visual-memory variants are introduced alongside their results in \cref{sec:supp-visual}. Retrieval, cross-day abstraction and faithfulness protocols accompany their results in \cref{sec:supp-results}; retention is described in \cref{sec:supp-ablation}.

\subsection{Memory Structure Analysis}
\label{sec:supp-behaviour-impl}

The results of this analysis are reported in \cref{sec:supp-behaviour,tab:supp-backbones}.

This section describes how we analyse the structure of the memories, so that the study can be repeated on any memory. Every step is computed from the operation log of a build. Except for the labelling of sentences, every step is a deterministic rule.

\noindent\textbf{Step 1: Memory.}
We replay the operation log and list every node with its parent, depth, children, time span, title and text. The interval of a node is the union of its span and the intervals of its children.

\noindent\textbf{Step 2: Fact Types.}
Every sentence of the memory is labelled with the kind of fact it states: action, speech, place or position,
object, time, pattern (a habit or a regularity), person, count, or other. Qwen3.8-27B receives the sentences 30 at
a time, with a one-line definition of each type, and gives each sentence the type most of its words serve. Embeddings are used to compute meaning similarity between nodes and refine the label assignment. We
labelled 1.46 million sentences, and shares are counted in tokens.
A node takes the type that holds most of its text. These labels are also used for \cref{fig:depth} in the main text.

\noindent\textbf{Step 3: Structures.}
A structure is an exact rule on a node and its children. We define 34 rules in nine families; the ones used in the paper are:
\begin{itemize}
\item \emph{moment} (leaf): a node without children, which describes one clip;
\item \emph{single fact}: a node with exactly one child, a leaf;
\item \emph{activity} (star): a node with at least two children, both without children themselves, that follow each other in time;
\item \emph{group of activities}: a node with at least two children that are themselves activities;
\item \emph{theme} (cross-day): a node whose children start on at least two days;
\item \emph{person node}: a node whose title names one person, with at least two children, and the majority of the text related to that person;
\item \emph{speech node}: at least two children, 80\% of them speech;
\item \emph{object node}: majority of the children are about objects;
\item \emph{pattern node}: at least 25\% of the text states patterns;
\item \emph{day node}: a child of the root titled with one day;
\item \emph{ordered}: at least three children, all in order of time.
\end{itemize}
For the topology figure of the main text, each node receives exactly one structure, the first rule it matches in a fixed order from the most specific to the most general (root, day, pattern, theme, person, speech, object, group of activities, activity, single fact, moment). A link of the figure counts the nodes of one structure that sit under a parent of another. The figure labels the root \emph{week}, pattern nodes \emph{habits}, speech nodes \emph{talk}, groups of activities \emph{episode} and single facts \emph{fact}.

\noindent\textbf{Step 4: Validation against chance.}
A structure could arise only from the number of nodes at each depth. We test this with a null model. Inside each memory organization, every node keeps its depth and its number of children and is assigned a random parent at the depth above. We draw 200 such organizations and count each structure in them. For \emph{ordered}, the null model permutes the children of a node. We call a structure \emph{robust} when it has at least 20 instances, is at least 1.5 times more frequent (or less frequent) than in the shuffled trees, with $|z|\geq 3$, in every build of the benchmark.

\noindent\textbf{Backbones.}
The structure comparison (\Cref{fig:structures} of the main text and \Cref{tab:supp-backbones}) uses one build per model on the same 35.5 hours of the week, with the same clips, challenges, categories and prompts. \Cref{fig:backbones} shows a trend, not a ranking of the backbones.

\section{Experiments}
\label{sec:supp-exp}

\begin{figure*}[!t]
\centering
\includegraphics[width=\textwidth]{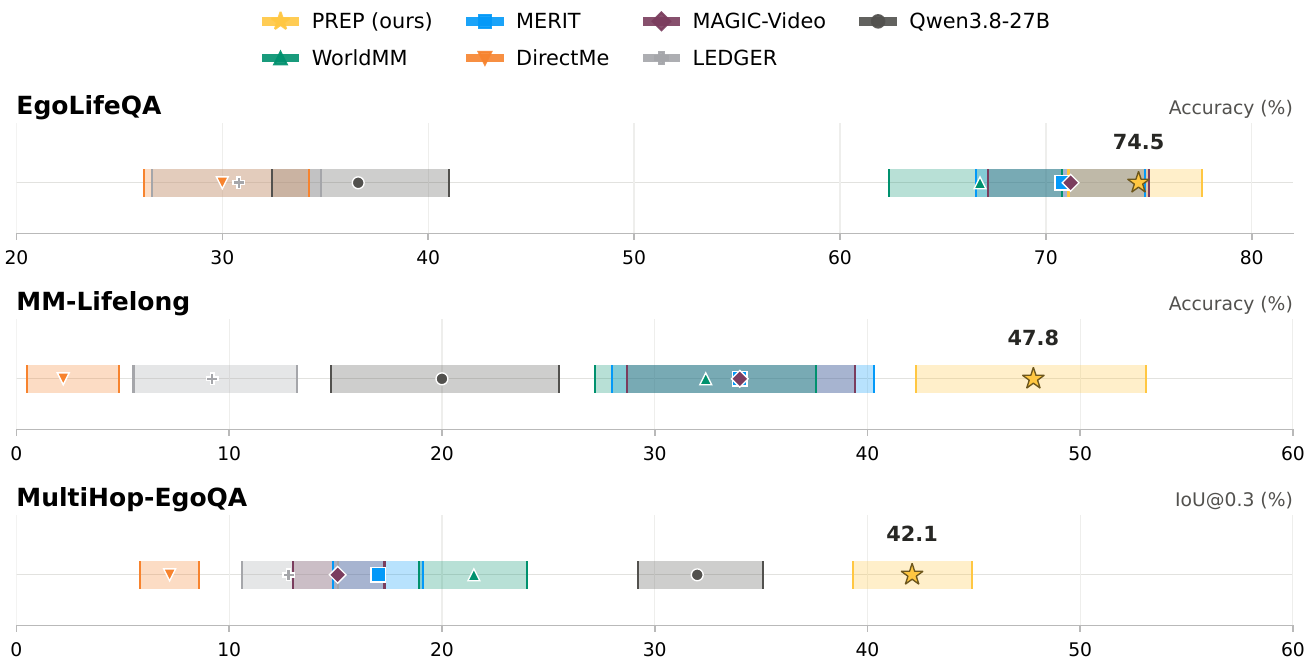}
\caption{\textbf{Scores and 95\% confidence intervals.} All methods share one horizontal line in each panel. Markers show scores; translucent bands and endpoint ticks show the intervals. Marker shapes distinguish methods, including coincident scores. Colors match the retrieval figure; the backbone and LEDGER use gray tones. The interval of \ours{} overlaps those of MERIT and MAGIC-Video on EgoLifeQA and is separated from every baseline on MM-Lifelong and MultiHop-EgoQA. Significance tests use paired bootstrap differences (\cref{sec:supp-ci-impl}).}
\label{fig:supp-ci}
\label{tab:supp-ci}
\end{figure*}

\subsection{Confidence Intervals and Repeated Runs}
\label{sec:supp-ci}

\noindent\textbf{Intervals.}
We use the paired bootstrap protocol in \cref{sec:supp-ci-impl}.
\Cref{fig:supp-ci} shows the scores and intervals. On MM-Lifelong the lead of \ours{} over the best baseline is 13.8 points, with interval $[8.1, 19.3]$. On MultiHop-EgoQA the lead over the best other method with our backbone, the backbone alone, is 10.1 points $[7.4, 13.1]$, and the answer score is 4.52 $[4.25, 4.78]$ against 3.73 $[3.50, 3.97]$ for MAGIC-Video, the best memory method. On UCS-Bench \ours{} scores 53.6 $[52.2, 54.8]$ and on Beyond3D 62.5 $[60.4, 64.6]$. The differences on MM-Lifelong and MultiHop-EgoQA have $p<0.001$. On EgoLifeQA \ours{} has the highest score and its interval overlaps those of MERIT and MAGIC-Video, as stated in the main text.

\noindent\textbf{Repeated runs.}
The results are stable under repetition, both of the answering and of the whole build.
\emph{Answering.} The five answering passes on the same memory give 75.2, 73.6, 74.4, 73.6 and 75.6 on EgoLifeQA and 49.0, 47.5, 47.3, 48.3 and 46.8 on MM-Lifelong (standard deviation 0.9 on both).
\emph{Building.} We rebuilt the memory of the week three more times from scratch. The rebuilt memories score 74.8, 75.2 and 74.6 on EgoLifeQA and 47.3, 48.5 and 49.0 on MM-Lifelong. Three rebuilds of MultiHop-EgoQA give an IoU@0.3 of 43.5, 44.4 and 42.8. A rebuild moves a score by about one point, far less than the margins over the baselines on MM-Lifelong and MultiHop-EgoQA.

\noindent\textbf{Agreement between judges.}
On MM-Lifelong, the two judges give the same credit on 89 to 97\% of the questions of each method, rank the methods in the same order, and give \ours{} a similar lead (13.8 points with gpt-oss-120b, 14.5 with Qwen3.8-Flash-Next). On MultiHop-EgoQA, the order of the methods is also unchanged with the second judge.

\begin{table*}[!t]
\centering
\small
\setlength{\tabcolsep}{3pt}
\caption{\textbf{Episodic benchmarks by category.} For MM-Lifelong, Other pools four types with fewer than 15
questions each (33 questions). All metrics are percentages except the answer score. Best in \textbf{bold},
second \underline{underlined}. $^\dagger$~Reported by the authors; the MM-Lifelong value uses a GPT-5 judge.}
\label{tab:episodic}
\resizebox{\textwidth}{!}{%
\begin{tabular}{l cccccc cccccccc ccccc}
\toprule
 & \multicolumn{6}{c}{EgoLifeQA}
 & \multicolumn{8}{c}{MM-Lifelong}
 & \multicolumn{5}{c}{MultiHop-EgoQA} \\
\cmidrule(lr){2-7}\cmidrule(lr){8-15}\cmidrule(lr){16-20}
Method
 & \rotatebox{90}{EntityLog} & \rotatebox{90}{EventRecall}
 & \rotatebox{90}{HabitInsight} & \rotatebox{90}{RelationMap}
 & \rotatebox{90}{TaskMaster} & \rotatebox{90}{Avg.}
 & \rotatebox{90}{Counting} & \rotatebox{90}{Temporal}
 & \rotatebox{90}{EntityRecog.} & \rotatebox{90}{Causal}
 & \rotatebox{90}{EventRecog.} & \rotatebox{90}{SpeechRecall}
 & \rotatebox{90}{Other} & \rotatebox{90}{Avg.}
 & \rotatebox{90}{IoU@0.3} & \rotatebox{90}{mIoU}
 & \rotatebox{90}{mIoP} & \rotatebox{90}{mIoG}
 & \rotatebox{90}{Score (0--10)} \\
\midrule
\multicolumn{20}{l}{\textit{General MLLMs}} \\
Cambrian-S-7B
 & 34.4 & 33.3 & 29.5 & 27.2 & 38.1 & 32.2
 & 16.9 & 15.3 & 13.0 & 14.7 & 5.9 & 3.3 & 1.5 & 11.5
 & 6.2 & 7.5 & 10.6 & 32.0 & 1.24 \\
VideoChat-Flash-7B
 & 39.2 & 42.9 & 26.2 & 30.4 & 38.1 & 36.2
 & 14.4 & 19.4 & \underline{26.1} & 8.8 & 5.9 & 3.3 & 15.2 & 14.8
 & 6.0 & 7.8 & 11.2 & 33.6 & 1.96 \\
InternVideo2.5-8B
 & 28.0 & 40.5 & 42.6 & 34.4 & 34.9 & 35.4
 & 18.6 & 13.9 & 17.4 & 11.8 & 5.9 & 3.3 & 21.2 & 15.3
 & 7.3 & 8.0 & 11.5 & 33.6 & 2.82 \\
Nemotron3.5-120B
 & 30.4 & 38.1 & 44.3 & 38.4 & 31.7 & 36.2
 & 11.9 & 18.1 & 10.9 & 5.9 & 11.8 & 0.0 & 15.2 & 12.0
 & 20.8 & 16.3 & 26.9 & 28.9 & 3.36 \\
Gemma4-31B-IT
 & 38.4 & 35.7 & 39.3 & 31.2 & 36.5 & 35.8
 & 7.6 & 15.3 & 13.0 & 14.7 & 5.9 & 0.0 & 13.6 & 10.5
 & \underline{38.8} & \underline{28.2} & \underline{41.2} & 45.9 & \underline{3.76} \\
Qwen3.8-27B
 & 35.2 & 40.5 & 37.7 & 36.0 & 31.7 & 36.6
 & 12.7 & 31.9 & 19.6 & 14.7 & 26.5 & 3.3 & 27.3 & 20.0
 & 32.0 & 24.7 & 36.3 & 44.4 & 3.64 \\
\midrule
\multicolumn{20}{l}{\textit{Memory Methods}} \\
GEB$^\dagger$
 & -- & -- & -- & -- & -- & \underline{72.0}
 & -- & -- & -- & -- & -- & -- & -- & 36.8
 & 25.9 & 20.5 & -- & -- & 3.11 \\
MERIT
 & \underline{71.2} & 68.3 & 75.4 & 72.0 & 68.3 & 70.8
 & \underline{28.0} & 36.1 & \underline{26.1} & \underline{38.2}
 & 41.2 & 43.3 & \underline{37.9} & \underline{34.0}
 & 17.0 & 14.5 & 19.2 & 42.9 & 3.19 \\
MAGIC-Video
 & \underline{71.2} & \textbf{70.6} & \underline{77.0}
 & 67.2 & \underline{74.6} & 71.2
 & 26.6 & \underline{44.0} & 23.9 & 29.4
 & \underline{48.0} & \underline{55.6} & 28.8 & \underline{34.0}
 & 15.1 & 13.7 & 16.6 & \underline{47.2} & 3.73 \\
WorldMM
 & 58.4 & 61.9 & 75.4 & \underline{73.6} & 71.4 & 66.8
 & 23.7 & 43.5 & 21.7 & 27.5 & 33.3 & 53.3 & 35.9 & 32.4
 & 21.5 & 17.5 & 23.7 & 46.3 & 3.22 \\
DirectMe
 & 22.4 & 27.0 & 32.8 & 37.6 & 33.3 & 30.0
 & 1.7 & 2.8 & 4.3 & 5.9 & 0.0 & 0.0 & 1.5 & 2.2
 & 7.2 & 7.5 & 17.2 & 18.3 & 3.08 \\
LEDGER
 & 28.8 & 34.1 & 37.7 & 28.0 & 27.0 & 30.8
 & 11.0 & 11.1 & 8.7 & 8.8 & 0.0 & 0.0 & 13.6 & 9.2
 & 12.8 & 11.2 & 23.6 & 26.2 & 2.56 \\
\midrule
\textit{Ours}
 & \textbf{71.8} & \underline{70.2} & \textbf{82.0}
 & \textbf{75.4} & \textbf{79.4} & \textbf{74.5}
 & \textbf{37.1} & \textbf{53.3} & \textbf{43.5} & \textbf{54.7}
 & \textbf{68.2} & \textbf{63.3} & \textbf{42.4}
 & \textbf{47.8}
 & \textbf{42.1} & \textbf{30.8} & \textbf{44.6}
 & \textbf{48.3} & \textbf{4.52} \\
\bottomrule
\end{tabular}%
}
\end{table*}

\subsection{Results by Question Type}
\label{sec:supp-categories}

Tables~\ref{tab:episodic} and~\ref{tab:spatial} break the scores of \cref{tab:main} down by question type, for all methods of the main table. The settings are those of \cref{sec:main}. GEB reports totals only.

\noindent\textbf{Episodic benchmarks (Table~\ref{tab:episodic}).} On EgoLifeQA, \ours{} is first in four of the five categories. The largest margin is on HabitInsight (82.0 against 77.0), the questions about what people usually do: these are the questions that a generalization in the memory answers directly. On EventRecall, MAGIC-Video is 0.4 points ahead. On MM-Lifelong, \ours{} is first in every category. The margins are largest where the answer must be assembled from several moments: 20.2 points on event recognition, 17.4 on entity recognition and 16.5 on causal questions. On MultiHop-EgoQA, \ours{} is first on every grounding metric and on the answer score. Every other memory method localizes the evidence worse than the base model without memory, while \ours{} improves on it by 10.1 points of IoU@0.3. The general video models stay below 45\% on EgoLifeQA and below 21\% on MM-Lifelong: the frames given with the question are not enough, and these questions need a memory.

\begin{table*}[!t]
\centering
\small
\setlength{\tabcolsep}{3pt}
\caption{\textbf{Spatial benchmarks by question type.} Accuracy (\%) on the 515 evaluated UCS-Bench videos and on
the eight Beyond3D question types (8{,}800 scored questions), whose average is the official score. Best in \textbf{bold}, second
\underline{underlined}. DirectMe on UCS-Bench is our run of its released code (42.1). Its authors report 50.5, the value underlined in the main table.}
\label{tab:spatial}
\resizebox{\textwidth}{!}{%
\begin{tabular}{l ccccc ccccccccc}
\toprule
 & \multicolumn{5}{c}{UCS-Bench}
 & \multicolumn{9}{c}{Beyond3D} \\
\cmidrule(lr){2-6}\cmidrule(lr){7-15}
Method
 & \rotatebox{90}{Obj. rec., quant.}
 & \rotatebox{90}{Pos., orient.}
 & \rotatebox{90}{Traj., movem.}
 & \rotatebox{90}{Proxim., reachab.}
 & \rotatebox{90}{All}
 & \rotatebox{90}{Visible now}
 & \rotatebox{90}{Last seen}
 & \rotatebox{90}{Last placed}
 & \rotatebox{90}{Nearest fixt.}
 & \rotatebox{90}{Camera: dir.}
 & \rotatebox{90}{Camera: dist.}
 & \rotatebox{90}{Object: dir.}
 & \rotatebox{90}{Object: dist.}
 & \rotatebox{90}{Avg.} \\
\midrule
\multicolumn{15}{l}{\textit{General MLLMs}} \\
Cambrian-S-7B
 & 33.4 & 37.2 & 41.0 & 49.3 & 38.9
 & 54.2 & 20.9 & 20.4 & 51.5 & 27.5 & 33.3 & 34.9 & 36.8 & 34.9 \\
VideoChat-Flash-7B
 & 40.0 & 38.3 & 43.5 & 49.4 & 42.0
 & 53.9 & 22.3 & 23.0 & 51.8 & 29.2 & 33.0 & 33.8 & 34.2 & 35.2 \\
InternVideo2.5-8B
 & 39.0 & 38.3 & 42.5 & 50.5 & 41.7
 & 58.9 & 22.2 & 23.7 & 52.1 & 27.8 & 39.2 & 33.2 & 34.6 & 36.5 \\
Nemotron3.5-120B
 & 38.6 & 41.1 & 51.6 & 52.6 & 44.6
 & -- & -- & -- & -- & -- & -- & -- & -- & 35.9 \\
Gemma4-31B-IT
 & 43.8 & \textbf{48.8} & \underline{55.8} & 57.7 & 50.2
 & -- & -- & -- & -- & -- & -- & -- & -- & 22.0 \\
Qwen3.8-27B
 & \underline{45.7} & 46.1 & 55.6
 & \underline{59.2} & \underline{50.4}
 & 75.3 & 40.4 & 42.4 & 54.4 & 49.3 & \underline{52.9}
 & 50.3 & 52.6 & 52.2 \\
\midrule
\multicolumn{15}{l}{\textit{Memory Methods}} \\
MERIT
 & 41.3 & 35.1 & 49.0 & 48.7 & 42.8
 & 84.8 & 36.1 & 36.5 & \underline{59.7} & 48.3 & 50.1
 & 50.7 & 53.0 & 52.4 \\
MAGIC-Video
 & 44.5 & 43.5 & 55.1 & 55.8 & 48.6
 & \underline{85.5} & \underline{41.2} & \underline{48.1}
 & 58.6 & \underline{53.4} & 49.5
 & \underline{52.1} & \underline{56.5} & \underline{55.6} \\
WorldMM
 & 41.1 & 39.8 & 49.6 & 50.0 & 44.2
 & 79.2 & 29.1 & 32.3 & 52.8 & 43.9 & 48.4
 & 45.9 & 47.6 & 47.4 \\
DirectMe
 & 38.6 & 39.8 & 42.9 & 51.3 & 42.1
 & 63.1 & 17.2 & 21.4 & 34.4 & 31.3 & 40.0
 & 35.9 & 35.3 & 34.8 \\
LEDGER
 & -- & -- & -- & -- & --
 & 61.4 & 20.3 & 21.8 & 33.1 & 32.2 & 37.1
 & 35.1 & 34.2 & 34.4 \\
\midrule
\textit{Ours}
 & \textbf{50.1} & \underline{48.4} & \textbf{58.7}
 & \textbf{59.7} & \textbf{53.6}
 & \textbf{89.3} & \textbf{61.4} & \textbf{60.3}
 & \textbf{60.9} & \textbf{55.2} & \textbf{53.3}
 & \textbf{56.3} & \textbf{62.6} & \textbf{62.5} \\
\bottomrule
\end{tabular}%
}
\end{table*}

\noindent\textbf{Spatial benchmarks (Table~\ref{tab:spatial}).} On UCS-Bench, \ours{} has the best overall score and is first in three of the four categories. On position and orientation it is within 0.4 points of the best model. Among the methods with our backbone, \ours{} is first in all four categories, and it is the only memory method that improves on the backbone in each of them. On Beyond3D, \ours{} is first in all eight question types. The margins are largest on the questions that need the past: 20.2 points on where an object was last seen and 12.2 on where it was last placed. They are smallest on the questions that the current frames can answer, such as the distance from the camera (0.4 points). The memory therefore helps where a memory is needed and costs nothing where it is not.

\subsection{Statistics of the Predictive Branch}
\label{sec:supp-predictive}

The prediction protocol is described in \cref{sec:supp-predictive-impl}.

\noindent\textbf{Challenges.}
Three challenges per clip give 18{,}750 challenges on the EgoLife week (6{,}250 clips), 60{,}612 on UCS-Bench and 6{,}204 on Beyond3D. Moreover, a challenge needs no new text when the Predictor answers it correctly, or when the Updater's check finds that the memory already states the answer. \Cref{tab:supp-pred} gives both shares. On UCS-Bench, half of the challenges (51.1\%) need no new text. On Beyond3D 35.3\% and on the EgoLife week 22.2\%. The week is the hardest case. A challenge there asks about a new conversation or event in an open-ended life recording, while a challenge on UCS-Bench or Beyond3D asks about a scene the memory has already mapped. The remaining challenges are the informative ones, as they point the Updater at what the memory lacks.

\begin{table}[!htb]
\centering
\small
\setlength{\tabcolsep}{4pt}
\caption{\textbf{Challenges the memory already covers}: the share of challenges that the previous memory $M_{t-1}$ predicts correctly (Predicted) or states explicitly (Stated), so no memory update is needed. Totals are computed before rounding.}
\label{tab:supp-pred}
\begin{tabular}{lcccc}
\toprule
 & Challenges & Predicted & Stated & Total \\
\midrule
EgoLife week & 18{,}750 & 4.1\% & 18.2\% & 22.2\% \\
UCS-Bench & 60{,}612 & 23.0\% & 28.1\% & 51.1\% \\
Beyond3D & 6{,}204 & 16.5\% & 18.8\% & 35.3\% \\
\bottomrule
\end{tabular}
\end{table}

\noindent\textbf{Change over time.}
The memory predicts more as it sees more. Inside a video, the share of correct predictions starts at 0\% on the first clip, when the memory is empty, and rises to 28.6\% late in the video on UCS-Bench and to 28.3\% on Beyond3D. On the EgoLife week, the share of challenges whose answer the memory already states is 18.2\%, between 11.5\% and 27.0\% by day.

\noindent\textbf{Retaining memory updates.}
We re-asked 1{,}000 failed challenges to the Answerer. Before the update of their clip, it answers 13.5\% of them. Right after the update it answers 76.2\%. At the end of the week, up to seven days and thousands of updates later, it answers 77.6\%. What the memory learns from a failure is kept for the rest of the stream. The gain holds in every category: $+81.7$ points on speech, $+68.1$ on people's actions, $+65.0$ on the wearer's actions, $+58.4$ on temporal order, $+52.1$ on inference, $+49.0$ on objects and $+43.4$ on spatial questions.

\subsection{Streaming Latency and Throughput}
\label{sec:supp-live-results}

The live and batched execution settings are described in \cref{sec:supp-streaming}.

\noindent\textbf{Live stream: one clip at a time.}
We ran this setting in real time on the EgoLife week, on one H200 (\cref{tab:supp-live}). The predictive branch takes 8.9 seconds and the update 14.4 seconds. The memory is updated on average after 23.3s, which is less than one clip. The delay does not grow over time, so \ours{} keeps pace with the stream.

\begin{table}[!htb]
\centering
\small
\setlength{\tabcolsep}{5pt}
\caption{\textbf{Time per call on a live stream.} Average and max on the EgoLife week in real time, in seconds, on a single H200. Each row in italics is the sum of its calls. The Reorganizer and the Answerer do not delay the update of the memory.}
\label{tab:supp-live}
\begin{tabular}{lcc}
\toprule
Call & Average & Maxima \\
\midrule
Challenger & 6.7 & 8.2 \\
Predictor & 1.1 & 1.9 \\
Judge & 1.1 & 1.6 \\
\textit{Predictive branch} & \textit{8.9} & \textit{11.7} \\
\midrule
Updater (check) & 9.4 & 11.6 \\
Updater (write) & 5.0 & 7.8 \\
\textit{Update} & \textit{14.4} & \textit{19.4} \\
\midrule
End of clip to updated memory & 23.3 & 31.1 \\
\midrule
Reorganizer (background) & 9.1 & -- \\
Answerer (one question, wall-clock) & 19.8 & -- \\
\bottomrule
\end{tabular}
\end{table}

\noindent\textbf{Cost in latency.}
Batching delays the update of the memory, because a batch waits for its later clips. With batches of six the memory is updated 113 seconds after a clip ends (median), against 23.3 seconds with one clip at a time. The gain is throughput: the week is built about six times faster than real time on one GPU.

\subsection{Cost of Memory Building}
\label{sec:supp-cost}

Module settings and baseline implementations are given in \cref{sec:supp-system,sec:supp-baselines}.

\noindent\textbf{Total cost of \ours{}.}
Building the memory of the 51.95-hour week takes 8.9 GPU-hours on one H200, that is 0.17 GPU-hours per hour of video, about six times faster than real time. This is the complete cost: decoding, the three calls of the predictive branch, the updates and the Reorganizer. On benchmarks without captions it also includes captioning, which is part of the Challenger call. The build of MultiHop-EgoQA takes 2.7 GPU-hours for 18 hours of video, and UCS-Bench 35.8 GPU-hours for 167 hours. Transcription with Whisper large-v3-turbo, where it is needed, uses a model about thirty times smaller than the backbone.

\noindent\textbf{Comparison with baselines.}
\Cref{tab:supp-cost} puts the three stages of a memory side by side: building, storing and answering. The build cost of \ours{} lies within the range of the baselines: below WorldMM, comparable to MAGIC-Video. MERIT builds a flat index at a lower cost and moves the work to the question: it needs 25.8 GPU-seconds per answer, against 14.0 for \ours{}. The cost of a build is paid once, the cost of an answer is paid for every question. \ours{} matches the lowest answering cost among the memory methods that read video, and its memory is 5 to 18 times smaller in text and over 1{,}000 times smaller in total, because it keeps no video. The baselines must keep 1 to 2\,GB of video per hour to answer.

\begin{table}[!htb]
\centering
\small
\setlength{\tabcolsep}{3pt}
\caption{\textbf{Cost of the three stages of a memory.} Build: GPU-hours per hour of video, range over the benchmarks; for \ours{}, the EgoLife week, MultiHop-EgoQA and UCS-Bench. Size: per hour of EgoLife video. Answer: GPU-seconds per question. In our runs DirectMe builds its graph in stream order and answers each question from the graph of the frames before it (its released evaluation builds the graph from the whole video). We did not time the build apart, so it is counted in the answer cost.}
\label{tab:supp-cost}
\resizebox{\columnwidth}{!}{%
\begin{tabular}{lcccc}
\toprule
 & Build & \multicolumn{2}{c}{Size (MB/h)} & Answer \\
\cmidrule(lr){3-4}
Method & (GPU-h/h) & text & total & (GPU-s) \\
\midrule
MERIT & 0.04--0.08 & 5.0 & 1{,}985 & 25.8 \\
MAGIC-Video & 0.11--0.24 & 13.2 & 1{,}993 & 14.0 \\
WorldMM & 0.48--1.87 & 9.1 & 1{,}989 & 73.2 \\
DirectMe & -- & 17.05 & 1{,}033 & 79.1 \\
\midrule
\ours{} & 0.15--0.21 & \textbf{0.93} & \textbf{0.93} & 14.0 \\
\bottomrule
\end{tabular}}
\end{table}

\noindent\textbf{Clip length vs. cost.}
With 30-second clips the week costs 8.9 GPU-hours and scores 74.5 on EgoLifeQA. With 5-second clips it costs 38.7 GPU-hours and scores 73.1. Longer clips are four times cheaper and as accurate, so the cost of \ours{} can be set by one parameter.

\subsection{Additional Ablations}
\label{sec:supp-ablation}

\begin{figure}[!htb]
  \centering
  \includegraphics[width=\columnwidth]{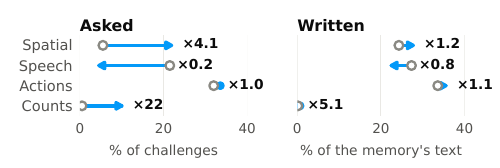}
  \caption{\textbf{Categories shift what is stored.} Left: share of the challenges asked about each kind of fact.
  Right: share of the memory's text that states each kind of fact, every sentence labelled with its main kind. Circles: without categories.}
  \label{fig:categories}
\end{figure}

\noindent\textbf{Question Categories.}
Category initialization is optional, serving as a personalization mechanism rather than a requirement.
Removing the benchmark-specific categories preserves accuracy ($-0.1$ points on EgoLifeQA, $-1.3$ on MM-Lifelong) and memory size.
As shown in Fig.~\ref{fig:categories}, combining the categories of several benchmarks increases spatial challenges fourfold, raising the share of position-related memory from 24\% to 29\%. Thus, categories allow users to steer memory toward their interests without retraining.

\noindent\textbf{Iterative search.}
On EgoLifeQA the full method scores 74.5. With a single retrieval and no further search it scores 68.8, with about six times fewer prompt tokens. With a single retrieval \ours{} still leads on MM-Lifelong (30.0, against 26.3 and 23.3 for the best baselines in the same setting). The iterative search is thus a choice between cost and accuracy, not a requirement for the lead.

\noindent\textbf{Memory retention.}
EgoLifeQA asks each question soon after its event. To test what a memory keeps, we ask the same questions again
1 hour and 24 hours after their own time and at the end of the week. We keep only the 154 of 500 questions whose
meaning does not depend on when they are asked: no word such as just, now, last, today, yesterday, this or a
present tense, and no link to the last occurrence before the asking time. The end-of-week delay ranges from minutes
to 6.4 days (3.5 on average). Each line is a least-squares fit of the change in each answer ($+100$, $-100$ or 0
against its own time) to its delay in days, with a 95\% interval from resampling the questions.

We ask the EgoLifeQA questions that do not depend on the asking time again at the end of the week, when the memory holds up to seven more days. With the tree, accuracy at the end of the week is level with the asking time, and its 95\% interval $[-9.6, +1.8]$ includes zero. With the same content but no polymorphic structure, it is 16.2 points lower ($[-22.8, -9.6]$). The generalizations near the root keep old facts findable as the memory grows.

\noindent\textbf{Cheaper answering.}
A memory built with challenges is easier to search. Compared with a build without challenges, the Answerer makes fewer calls to the memory per question (0.40 fewer on EgoLifeQA, 0.76 fewer on MM-Lifelong). Compared with a memory without a tree, an answer on EgoLifeQA uses 7{,}400 fewer prompt tokens and 28 seconds less.

\subsection{Visual Memory}
\label{sec:supp-visual}

\label{sec:supp-visual-impl}

\ours{} stores text only. We tested whether stored images would help, in two forms.

\noindent\textbf{Keyframes in the memory.}
The Updater keeps one image with each node it writes: a frame of the clip, or a crop of it that shows the fact. The Answerer sees the images of the nodes it reads, at most 16 per question.

\noindent\textbf{Frames on request.}
The Answerer gets a tool that returns the frames of a time range it names, so it can look at the video behind a node.

Images remain a compatible extension: a node can carry a keyframe without any change to the rest of the method, for applications that need exact visual detail.

\noindent\textbf{Results.}
On MultiHop-EgoQA, 95\% of the new nodes keep an image, and 68\% of the images are crops.
Frames help to place an event precisely in time. On MultiHop-EgoQA, which scores the time intervals of the evidence, both forms raise IoU@0.3 by 1.8 points on the same build, and keyframes raise the answer score by 0.15 overall. These gains are within the confidence interval. On EgoLifeQA, where the questions ask what happened, who said what and what people usually do, keyframes do not change the result. The text memory already holds what these questions need, and the frames the benchmark gives with the question cover the present.
Keyframes multiply the storage by 19 (17\,MB more per hour) for a gain that stays within the interval. Text is therefore the default of \ours{}.


\subsection{Memory Structure and Stability}
\label{sec:supp-behaviour}

We analyse the learned trees using the procedure in \cref{sec:supp-behaviour-impl}. Their semantic organization emerges within the shared rooting and generality rules, without predefined node types, branching or depth.

\noindent\textbf{Validation against chance.}
On the EgoLife week, groups of activities are 12 to 44 times more frequent than chance, ordered children 10 to 30 times, person nodes 12 to 49 times, and speech nodes 3 to 7 times. The structures are thus a property of how the model organizes the stream, not of the size of the tree.

\noindent\textbf{Backbones.}
\Cref{tab:supp-backbones} gives the statistics behind the figure of the main text. The differences are large: Gemma builds trees more than twice as deep as Qwen and 171 nodes that span several days, against 2. Gemini opens the week with 5 nodes, Qwen with 486.

\begin{table}[!htb]
\centering
\small
\setlength{\tabcolsep}{3pt}
\caption{\textbf{Memory structure per backbone}, same 35.5 hours of the EgoLife week, same settings.}
\label{tab:supp-backbones}
\resizebox{\columnwidth}{!}{%
\begin{tabular}{lcccc}
\toprule
 & Qwen & Gemini & Gemma & Nemotron \\
\midrule
Nodes per video hour & 248 & 140 & 278 & 211 \\
Depth, maximum & 6 & 6 & 14 & 14 \\
Depth, mean & 2.7 & 3.9 & 6.3 & 2.6 \\
Children of the root & 486 & 5 & 7 & 419 \\
Nodes spanning several days & 2 & 2 & 171 & 6 \\
Person and place nodes & 1 & 0 & 54 & 3 \\
Titles with a clock time (\%) & 96 & 100 & 3 & 77 \\
Characters per node body & 489 & 1{,}425 & 169 & 354 \\
\bottomrule
\end{tabular}}
\end{table}

\noindent\textbf{Repeated runs.}
The modules sample their output, so two builds of the same video never write the same nodes. The question is whether a backbone's structure is a habit of the model or an accident of one run. We built the memory of the week four times with Qwen3.8-27B, from scratch, and compared the four trees.
\begin{itemize}
\item \emph{Same structures.} Of the 15 structures tested against chance, 14 are on the same side of chance in all four builds. Activities cover 68 to 75\% of the nodes in every build, and ordered children are 12.6 to 14.6 times more frequent than chance.
\item \emph{Same mix.} The shares of the node kinds agree within one point: 83\% leaves, 2.1 to 2.6\% activities, 1.2 to 1.4\% person nodes. All four builds place seven day nodes under the week.
\item \emph{Same topology.} The matrices of links between structures, which the topology figure draws, have a cosine similarity of 0.993 to 0.998 between builds.
\item \emph{Same shape.} The distribution of local shapes differs between two builds by a Jensen--Shannon divergence of 0.039, close to the floor set by sampling (0.012). A build without the predictive branch is at 0.140 from them. For the distribution of depths, the builds are at 0.009 from each other and at 0.12 to 0.18 from other constructions.
\end{itemize}
The same holds on short videos. Across four builds of MultiHop-EgoQA, a clip gets 19.5 to 20.1 nodes at a mean depth of 2.82 to 2.84. A node that reappears in another build keeps a parent with the same meaning in 86.8\% of the cases, against 15.2\% for a shuffled tree.

\noindent\textbf{Why this matters.}
A repeated run changes the words of the memory and leaves its organization in place. The variation between runs of one model is an order of magnitude smaller than the variation between models in \cref{tab:supp-backbones}. The different structures that the backbones build are therefore reproducible properties of each model: each model has its own way of organizing experience, and \ours{} lets it appear because the prompts leave node types, branching and depth unspecified.

\subsection{Additional Results}
\label{sec:supp-results}

\label{sec:supp-additional-protocols}
Each evaluation below gives its setup followed by its results; common scoring conventions are described in \cref{sec:supp-bench}.

\noindent\textbf{Retrieval evaluation.}
Both measures count what the memory gave the answering model, never the frames of the benchmark. \emph{Within
60\,s} (the rule of MERIT): a hit is a retrieved item of at most 60 seconds within 60 seconds of the annotated
evidence. \emph{On the evidence}: A record line counts at its own time. A node counts with its whole span, so a node covering several
moments is not a hit even when it states the answer. \emph{Judged}: Qwen3.5-VL reads the question, the
correct answer and the retrieved text and frames, and says whether they state what the answer rests on. The first measure is strict on general
nodes and the second is not.

The retrieval judge agrees with a text-only judge (gpt-oss-120b) on 89.3\% of the questions. \ours{} generally leads under both the temporal and judged measures defined above.

\noindent\textbf{Memory abstractions.}
The benchmarks mostly ask about single moments. A memory kept over a week should also hold what is true across
days: a habit, a plan and what came of it, an object that keeps coming back. No benchmark on EgoLife tests this,
so we wrote 149 such questions about the EgoLifeQA week. We drafted candidates from the nodes of ten of our builds
that state a fact over several days, then checked each one against the dataset's raw annotations (dense captions
and transcripts). A question was kept only if the annotations support its answer on at least two different days. In 107 of the kept questions the answer combines facts from
different days, e.g., where the tart shells bought on the first evening were stored on DAY4 and on which evening
tarts were first made. In the other 42, one day is enough and the other days show that the fact recurs. All
questions are asked at the end of the week, answered in free text and graded by the judge used for MM-Lifelong.

Two controls bound the scores. Given only the annotation lines around the cited moments, the backbone answers
94.3\%, so the questions can be answered and their reference answers agree with the data. Given nothing, it answers
6.7\%, so they cannot be guessed. \ours{} answers 75.5\%. With the same backbone, MAGIC-Video answers 58.7\%,
WorldMM 55.4\% and MERIT 53.4\%. The questions were drafted from what our memories generalized, so they measure
whether the baselines hold the same cross-day knowledge, and are not meant to be used for fair comparison between \ours{} and the baselines.

\noindent\textbf{Faithfulness of the memory.}
Accuracy does not show whether the memory stores things that did not happen, so we check its text against the
video. We split 180 sampled nodes of the EgoLife week (60 from single batches, 60 above them, 60 by the
reorganizer) into 2,483 claims. Qwen3.8-27B (also the builder's model) and Gemma4-31B-IT label each claim from up
to 32 frames, the speech and the narration of the node's span.

Precision, the share of supported claims among
those supported or contradicted, is 91.5\% $[89.3, 93.5]$ with Gemma and 90.9\% $[88.2, 93.2]$ with Qwen, and 95.5\% on the 87.9\% of claims where both agree. Most errors are wrong actions or times. Precision shows no decline over the seven days (86 to 94\% by day with Gemma).

\noindent\textbf{Beyond Egocentric Video.}
Using only open-source models, we also applied \ours{} to Video-MME and LVBench, recording accuracy improvements of 14.7 percentage points on Video-MME Long and 21.7 on LVBench over the backbone alone. These results demonstrate applicability beyond egocentric video.

\section{Illustrative Memory Update}
\label{sec:supp-update-example}
The six steps below follow one 30-second observation from the prior memory to the committed update. This is a constructed example, not a recorded model run or an experimental result. Tool calls use readable shorthand.

\begin{quote}
\small
\begin{enumerate}
\renewcommand{\labelenumi}{\textbf{\arabic{enumi}.}}
\setlength{\itemsep}{0.8em}
\item \textbf{Start from the prior memory.}
Root 1, ``Home'', has children 2 (``Kitchen'') and 3 (``Repairs''). Node 2 has children 7 and 8; node 3 has child 9.
Node 7: ``Keys'', body ``09:00: Jake left the keys on the kitchen counter'', span 09:00--09:00:30.
Node 8: ``Mug'', body ``09:00: The blue mug is on the kitchen shelf'', same span.
Node 9: ``Repair contact'', body ``08:30: Shure coordinates the plumbing repair'', span 08:30--08:30:30.
Nodes 1--3 summarize their descendants.

The outline shown to the modules is: Breadth-first order: 1 (Home, 08:30--09:00:30, 2 children); 2 (Kitchen, 09:00--09:00:30, 2); 3 (Repairs, 08:30--08:30:30, 1); 7 (Keys, 09:00--09:00:30, 0); 8 (Mug, 09:00--09:00:30, 0); 9 (Repair contact, 08:30--08:30:30, 0). No bodies appear in the outline. The matched-node view supplies the bodies of 7--9.

\item \textbf{Read the new observation.}
Clip 09:10:00--09:10:30: 60 frames show Jake moving the keys from the counter into a drawer. The blue mug remains on the shelf.
Transcript: Tasha says, ``Shure, the plumber is coming at three this afternoon.''
Caption: ``Jake puts the keys into the kitchen drawer while Tasha tells Shure the plumber will arrive at 15:00. The blue mug is on the shelf.''

\item \textbf{Challenge, predict and judge.}
All three predictions use the same pre-clip memory. The Predictor sees neither the observation nor the reference answers.
\begin{enumerate}
\renewcommand{\labelenumii}{\textbf{(\arabic{enumii})}}
\setlength{\itemsep}{0.5em}
\item Question: ``Where did Jake put the keys?'' Category: object location.
\par Reference: ``In the kitchen drawer.'' Evidence: the frames show the transfer from counter to drawer.
\par Prediction: ``On the kitchen counter'' (node 7). Judge: \emph{wrong}, the prediction describes the earlier location.

\item Question: ``Whom did Tasha tell about the plumber, and at what time will the plumber arrive?'' Category: speech.
\par Reference: ``Shure; at 15:00 today.'' Evidence: the quoted transcript.
\par Prediction: ``Shure; the arrival time is unknown'' (node 9). Judge: \emph{partial}, the time is missing.

\item Question: ``Where is the blue mug?'' Category: object location.
\par Reference: ``On the kitchen shelf.'' Evidence: the mug is visible on the shelf.
\par Prediction: ``On the kitchen shelf'' (node 8). Judge: \emph{correct}.

\end{enumerate}

\item \textbf{Check whether the failed answers are already stored.}
The failed set contains challenges 1 and 2. Searches for ``keys drawer'' and ``plumber arrival'' and reads of nodes 7 and 9 find neither new fact. Thus neither failure is a retrieval miss, and no node is flagged for reorganization by this check. Challenge 3 needs no content update.

\item \textbf{Commit the memory edits.}
\begin{itemize}
\item \texttt{append(7,} ``09:10: Jake moved the keys into the kitchen drawer''\texttt{)}; node 7 retains its 09:00 statement and extends its span through 09:10:30.
\item \texttt{add(parent=3, title=} ``Plumber appointment''\texttt{, body=} ``09:10: Tasha told Shure the plumber will arrive today at 15:00''\texttt{)} creates node 10, span 09:10--09:10:30.
\item The correct prediction adds a dated ``seen again at 09:10'' line to node 8. The caption and transcript enter the record verbatim.
\end{itemize}

\item \textbf{Inspect the resulting memory.}
The only changes are the appended line in node 7, the dated recurrence in node 8, new node 10 under Repairs, the new record entry, and updated descendant spans. All earlier facts remain. A read of node 7 now gives both key locations with their times. The next outline includes node 10 and shows two children under Repairs.

No reorganization is needed in this example: the two new facts already fit the existing branches. The background Reorganizer may inspect the updated tree but commits no structural edit in this illustrative trace.
\end{enumerate}
\end{quote}

\section{Ethics Statement}
\label{sec:supp-ethics}

A memory of an egocentric stream records the wearer and also the people around them, who may not have agreed to it. A system that remembers a week of conversations could be misused to monitor others. We see four properties of \ours{} that reduce this risk, and three practices that a deployment should follow.

\emph{Properties.} (i) The memory holds text only. It stores no frames, faces or voices, and so no biometric data; a clip is discarded once it is processed. (ii) The whole system runs on one local GPU with open-weight models, so recordings and memories can remain private to the owner. (iii) The memory is plain text that a person can read. What it holds about someone can be inspected, corrected or removed on request, which is not possible for a memory of embeddings or of raw video. (iv) Plain-language categories steer what enters the tree, and the record can be switched off. A deployment could therefore leave out a kind of content, such as other people's speech, without retraining. We have not evaluated this setting.

\emph{Practices.} A deployment should make recording visible to bystanders and obtain consent where the law requires it, limit how long a memory is kept, and restrict access to the wearer.

All experiments use public datasets recorded with the consent of their participants. We collected no new recordings of people.

\end{document}